\documentclass[acmlarge]{acmart}

\usepackage{subfigure} 
\usepackage{multirow} 
\usepackage{algorithmic} 
\usepackage{algorithm} 
\AtBeginDocument{%
  }

\setcopyright{acmlicensed}
\copyrightyear{2025}
\acmYear{2025}
\acmDOI{10.1145/3729495}

\acmJournal{IMWUT}
\acmVolume{9}
\acmNumber{2}
\acmArticle{56}
\acmMonth{6}

\begin{document}

\title[Deconfounding Causal Inference through Two-Branch Framework with Early-Forking for Sensor-Based Cross-Domain HAR]{Deconfounding Causal Inference through Two-Branch Framework with Early-Forking for Sensor-Based Cross-Domain Activity Recognition}

\author{Di Xiong}
\email{221812013@njnu.edu.cn}
\orcid{0009-0009-6445-7863}
\affiliation{
        \department{School of Electrical and Automation Engineering}
	\institution{Nanjing Normal University}
	\city{Jiang Su}
	\country{China}}

\author{Lei Zhang}
\email{leizhang@njnu.edu.cn}
\orcid{0000-0001-8749-7459}
\affiliation{%
        \department{School of Electrical and Automation Engineering}
	\institution{Nanjing Normal University}
	\city{Naning}
	\state{Jiang Su}
	\country{China}}
\authornote{Corresponding author: Lei Zhang}

\author{Shuoyuan Wang}
\email{claytonwang0205@gmail.com}
\orcid{0000-0003-1795-4161}
\affiliation{
  \department{Department of Statistics and Data Science}
	\institution{Southern University of Science and Technology}
	\city{Shenzhen}
	\state{Guang Dong}
	\country{China}
}

\author{Dongzhou Cheng}
\email{230249457@seu.edu.cn}
\orcid{0000-0003-1575-6292}
\affiliation{
  \department{School of Cyber Science and Engineering}
	\institution{Southeast University}
	\city{Naning}
	\state{Jiang Su}
	\country{China}
}

\author{Wenbo Huang}
\email{230228503@seu.edu.cn}
\orcid{0000-0002-6664-1172}
\affiliation{
  \department{School of Computer Science and Engineering}
	\institution{Southeast University}
	\city{Naning}
	\state{Jiang Su}
	\country{China}
}

\renewcommand{\shortauthors}{Xiong et al.}


\begin{abstract}
 Recently, domain generalization (DG) has emerged as a promising solution to mitigate distribution-shift issue in sensor-based human activity recognition (HAR) scenario. However, most existing DG-based works have merely focused on modeling statistical dependence between sensor data and activity labels, neglecting the importance of intrinsic casual mechanism. Intuitively, every sensor input can be viewed as a mixture of causal (category-aware) and non-causal factors (domain-specific), where only the former affects activity classification judgment. In this paper, by casting such DG-based HAR as a casual inference problem, we propose a causality-inspired representation learning algorithm for cross-domain activity recognition. To this end, an early-forking two-branch framework is designed, where two separate branches are respectively responsible for learning casual and non-causal features, while an independence-based Hilbert-Schmidt Information Criterion is employed to implicitly disentangling them. Additionally, an inhomogeneous domain sampling strategy is designed to enhance disentanglement, while a category-aware domain perturbation layer is performed to prevent representation collapse. Extensive experiments on several public HAR benchmarks demonstrate that our causality-inspired approach significantly outperforms eleven related state-of-the-art baselines under cross-person, cross-dataset, and cross-position settings. Detailed ablation and visualizations analyses reveal underlying casual mechanism, indicating its effectiveness, efficiency, and universality in cross-domain activity recognition scenario.
\end{abstract}





\maketitle

\section{INTRODUCTION}
\subsection{Background}
\indent During recent years, the widespread use of portable wearable devices, such as smartwatches, fitness trackers, and smartphones, has made sensor-based human activity recognition (HAR) a hot research topic in ubiquitous computing community \cite{dang2020sensor,wang2019survey,wang2024optimization}. Sensor-based HAR aims to train machine learning models for recognizing human activities, through multivariate time series data collected from miniaturized inertial measurement units (e.g., accelerometer and magnetometer) attached to different body parts. HAR has played a crucial role in a large range of real-world scenarios, including personal fitness, sport tracking, health management, assisted living, gait analysis, and smart homes \cite{wang2019survey,bianchi2019iot}. In particular, due to an automatic feature extraction ability, deep learning models such as Convolutional Neural Networks (CNNs) \cite{ronao2016human}, Long-Short Term Memory (LSTM) \cite{bock2021improving}, and self-attention based Transformer \cite{mazzia2022action} have been widely adopted to improve the performance of HAR.

\subsection{Main Challenges}
In practice, most existing deep models are mainly built under the assumption that the training and test data follow the same distribution \cite{huang2022channel,lu2024diversify}. However, such assumption does not always hold true in real-world HAR deployment, where activity data usually tends to change with different distributions, since the sensor signals are typically affected by the biological factors of diverse users like their personal behavior styles, genders, and body shapes. For example, activity recognition models trained on the data from children are highly likely to fail while being tested on adults’activities. Standard training usually leads to entangled semantic and domain-specific features, which might vary drastically in new environments, and impede model generalization to a new domain \cite{qian2021latent,wang2022generalizing}. As a consequence, they would suffer from substantial performance degradation, which hinders their practical deployment for cross-domain activity recognition in a variety of real-world applications \cite{qin2022domain,qian2021latent}. To tackle such distribution-shift problem, Domain Generalization (DG) has increasingly gained a lot of attention, whose main goal is to learn domain-invariant representations from multiple source domains, that can generalize well on an unseen target domain. So far, there have existed many DG-based works that may be roughly categorized into three research lines: Data manipulation \cite{lu2024fixed,zhang2018mixup,zhoudomain}, Learning strategy\cite{sagawadistributionally,huang2020self}, and Representation learning \cite{ganin2016domain,du2021adarnn,lu2024diversify}. Although the past years have witnessed their promising results, DG still remains highly challenging, that is far from being solved in the context of HAR \cite{lu2024diversify,qian2021latent}. For instance, as introduced in a recent reference \cite{gulrajani2021search}, it turns out that without additional efforts, the vanilla empirical risk minimization (ERM) is even still able to substantially beat most of those existing DG-based alternative solutions that are elaborately designed, under rigorous evaluation protocols. Such observation furthermore highlights the demand of developing an effective DG-based algorithm, which is capable of capturing domain-invariant feature representations to mitigate such domain-shift problem in real-world HAR scenario.

\subsection{Research Motivation}
\begin{figure*}[!t]
	\centering
		\subfigure{
		\includegraphics[width=0.46\linewidth]{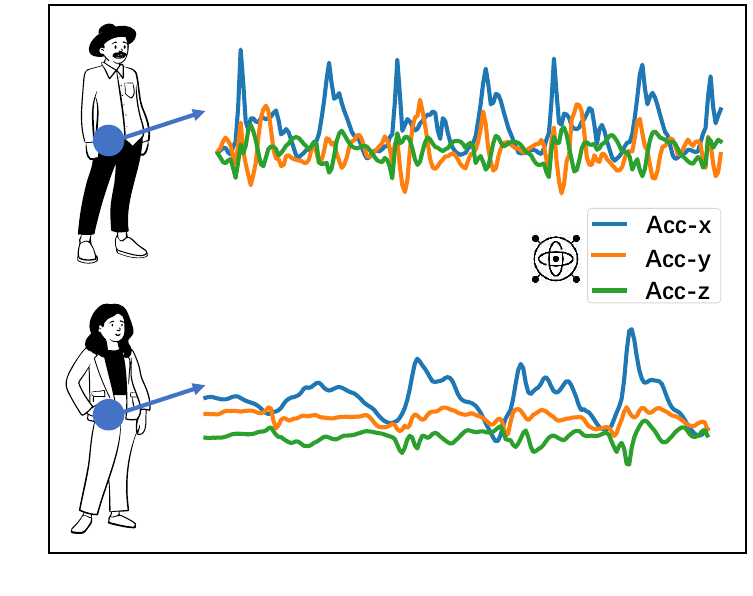}
		\label{fig:subfig_intr0_1}
	}
	\subfigure{
		\includegraphics[width=0.49\linewidth]{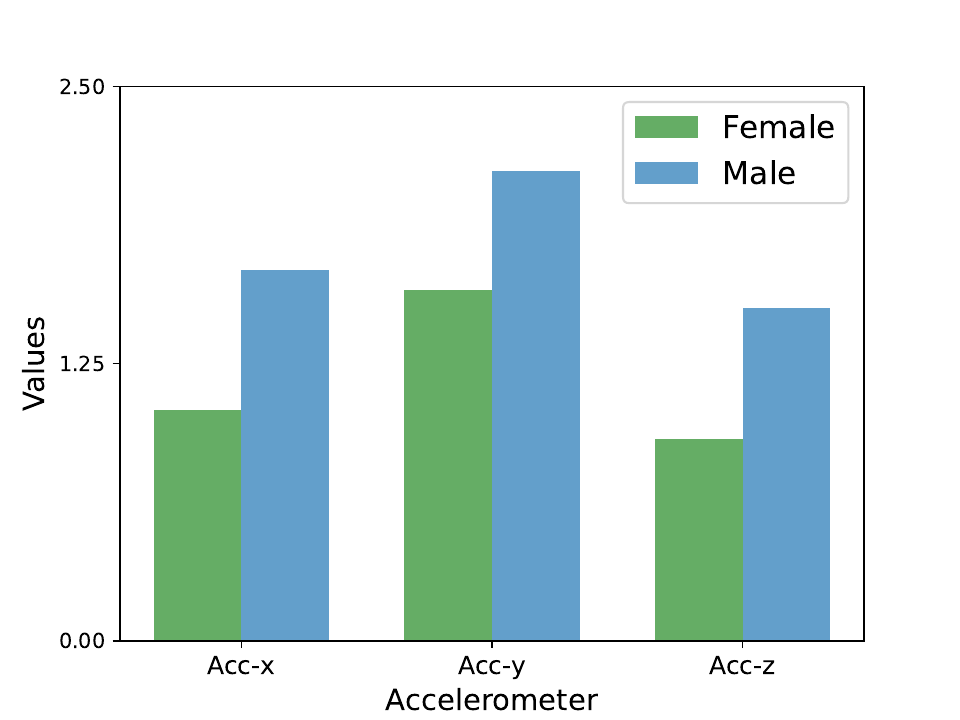}
		\label{fig:subfig_intro_2}
	}
	\caption{The \textbf{left subplot} illustrates the sensor readings of the “Running” activity collected from two subjects (i.e., an adult male and a female adult) in the USC-HAD dataset. Based on two different gender groups, the \textbf{right subplot} presents their averaged results by summing the absolute values of acceleration changes across all the three axes. It can be seen that adult males have greater amplitude of accelerometer output from all the three measurement directions than adult females. This implies that the sensor samples are often labeled in a compositional way, which make deep learning models also learn the feature representation in an entangled way. For a specific sensor sample labeled as “Running”, it is difficult to tell which features being an activity class, and which, being gender.}
	\vspace{-.1in}
	\label{fig_run_sensor}
\end{figure*}

To solve such domain-shift HAR problem, the current mainstream DG-based solutions are to model the statistical dependence between sensor data and activity labels, which intend to learn robust representations, independent of domain \cite{lu2024diversify,lu2024fixed,du2021adarnn,qian2021latent}. However, they have an obvious insufficiency of only modeling statistical dependence, while potentially ignoring the importance of intrinsic causal mechanism. In fact, the statistical dependence is only a superficial descriptor and always changes with the target distribution, which may fail to make the model generalize well. For instance, as shown in \figurename~\ref{fig_run_sensor}, in a cross-person activity recognition task, it is very likely that some male adults are running at a faster pace, showing high statistical dependence, which could easily mislead the model to make wrong predictions when the running pace slows down in target domain, e.g., females. After all, the statistical characteristics of sensor waveforms from running, instead of the subject gender plays a dominant role in cross-domain activity classification. Intuitively, different individuals should exhibit varying physiological characteristics, such as age, weight, and height, resulting in distinct activity data distributions. This variation in data distributions may be referred to as data heterogeneity. Another typical example is the public USC-HAD dataset \cite{zhang2012usc}, which involves a total of 14 subjects with different ages, weights, and heights. All subjects were above 21 years old, with weights in the range of 43 \textasciitilde 80 kg and heights in the range of 160 \textasciitilde 185 cm. Such insufficiency of merely modeling statistical dependence would lead to entangled semantic and domain-specific features. It has been recently argued \cite{peters2017elements} that such practices may be insufficient, and generalizing well outside the independent and identically distributed (IID) setting requires learning not mere statistical dependence between variables, but an underlying causal model \cite{bottou2013counterfactual,parascandolo2018learning,chen2024causal,spirtes2001causation,peters2017elements,scholkopf2012causal}.

\subsection{Main Contribution} \label{sec:our_approach}
In this paper, in order to enhance generalization capability on unseen domains, we propose to formulate such DG-based HAR task as a common casual inference problem \cite{atzmon2020causal,lv2022causality,chen2024causal}. To be specific, we treat the activity category-related information contained in sensor data as causal factors (e.g., ‘sensor signal waveform’), whose relationship associated with sematic label should be independent of domain. Instead, the domain-related information, e.g., ‘subject gender’ that is independent of activity category is generally viewed as non-causal factors. As illustrated in \figurename~\ref{fig_mt}, every raw sensor input X can be seen as a mixture of causal factors $X_c$ and non-causal factors $X_d$, which is considered to be generated through interventions on ‘Categories (C)’ and ‘Domains (D)’. In other words, the former is the main cause of the domain-invariant semantic (causal: $X_c$) and the latter is domain-specific features (non-causal: $X_d$), while only the former, i.e., sematic features have a causal effect on activity category label $y$. Such entanglement of causal and non-causal factors would lead to an undesired case. Given that an unexpected intervention is imposed over D at inference time, the non-causal features might also be encoded into the final classifier, which inevitably corrupt final activity prediction results. 

To alleviate above issue, we cast cross-domain activity recognition problem as a disentanglement procedure of the casual and non-causal features. Our main goal is to separate the causal factors $X_c$ from the non-casual ones $X_d$ within raw input $X$. Unfortunately, although these causal features are generally assumed to be independent of the non-causal ones, it is often difficult to achieve them by directly factorizing raw sensor input, since both causal and non-causal factors could not be explicitly formulated, which renders such causal inference highly challenging for sensor-based HAR. To put this intuitive idea into practice, we for the first time introduce a Causality-motivated Representation Learning algorithm to implement the factorization of causal/non-casual factors, which demonstrates stronger generalization ability in sensor-based cross-domain activity recognition scenario. A two-branch framework is leveraged to implicitly disentangle two kinds of features, where one branch is in charge of extracting the casual features, while the other branch is in charge of extracting the non-casual features. To the best of our knowledge, this paper is the first work that mainly concentrates on the sensor-based cross-domain activity recognition from a perspective of the disentanglement between causal and non-casual features. Overall, the main contributions of our work are three-fold:

\begin{figure}[!t]
	\centering
	\subfigure[]{
		\includegraphics[width=0.36\linewidth]{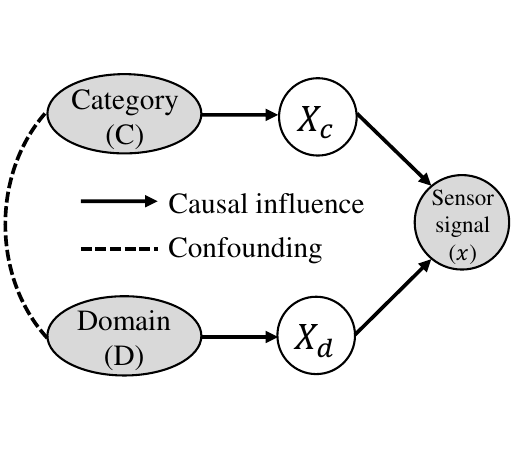}
		\label{fig:subfig_mt_1}
	}
	\subfigure[]{
		\includegraphics[width=0.58\linewidth]{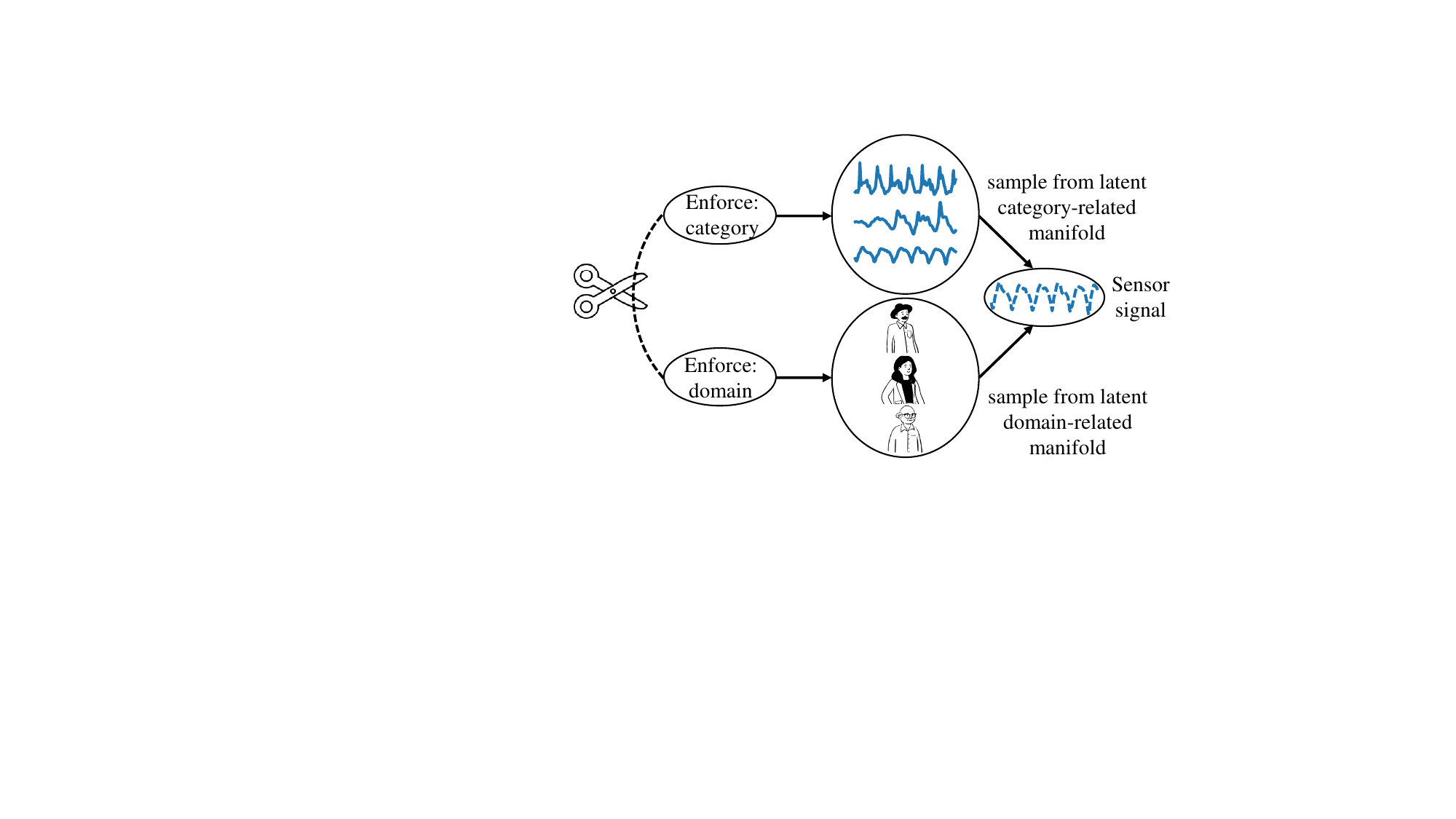}
		\label{fig:subfig_mt_2}
	}
	\caption{\textbf{(a)} A causal diagram illustrating the sensor signal generation for HAR. The two variables, "Category" and "Domain," produce causal features $X_c$ and non-causal features $X_d$ through causal influences, indicated by solid arrows. Both features then jointly generate the sensor signal. Dashed lines represent confounding between the two variables, indicating that they are not independent. \textbf{(b)} An intervention is applied to sever the connection between the two variables, a process referred to as disentangling causal and non-causal factors.}
    \vspace{-.3in}
	\label{fig_mt}
\end{figure}

\textbf{New perspective:} Most existing DG-based HAR works have caused entangled semantic and domain-specific features. This paper is the first time to point out the insufficiency of merely focusing on statistical dependence between sensor data and semantic label for sensor-based cross-domain activity recognition, and propose to excavate the underlying causal mechanisms from a new causality-inspired viewpoint. 

\textbf{Simple yet effective algorithm:} We cast such DG-based activity recognition as a common casual inference problem, and introduce a causality-inspired representation learning algorithm built on a two-branch structure with early-forking to disentangle casual and non-casual features. An inhomogeneous domain perturbation strategy is leveraged to enhance its generality capability, while a category-aware domain perturbation layer is performed to prevent representation collapse.

\textbf{Superior performance and extensibility:} Extensive experiments, ablation studies, and visualized results on four publicly available HAR benchmarks including DSADS, USC-HAD, PAMAP2, and UCI-HAR demonstrate the effectiveness, efficiency, and universality of our causality-inspired representation learning algorithm.

\section{RELATED WORK}
\subsection{Human Activity Recognition}
Human activity recognition (HAR) primarily targets at recognizing activities of daily living performed by different persons, which may be roughly categorized into two research streams according to data type: vision-based HAR and sensor-based HAR \cite{dang2020sensor}. The former mainly relies on activity data collected from camera or other optical device, which does not work well for HAR while a subject is beyond its coverage range \cite{kong2022human}. Moreover, it would suffer from severe privacy leaking problem, since sensitive personal data such as facial information might be accidentally released on cameras. During the past decade, the widespread adoption of low-cost, small-sized, and high-precision sensors has led to an increasing popularity of wearable and mobile devices, making sensor-based activity recognition an active research direction in the ubiquitous computing scenario \cite{huang2022channel,qian2021latent}. In this paper, we mainly focus on sensor-based HAR, due to its wide popularity of various wearable devices such as smartphones. Sensor-based HAR aims to train machine learning models by leveraging sensor data collected from multiple sensors (such as accelerometers, gyroscopes, and magnetometers) to identify a variety of human activities. Earlier researches have primarily concentrated on the use of traditional machine learning algorithms such as random forests, Naive Bayes, K-Nearest Neighbors (KNN), and Support Vector Machines (SVM) \cite{bao2004activity,khemchandani2016robust}, which heavily depend on time-consuming and laborious handcrafted feature design. Recently, deep learning models have attracted great research attention in this area. For instance, Zeng et al, Yang et al. and Ronao et al. have been the first to apply Convolutional Neural Networks (CNNs) to automatically capture discriminative feature representations from raw sensor signals for activity recognition tasks \cite{zeng2018understanding,yang2015deep,ronao2016human}. Ord{\'o}{\~n}ez et al. \cite{ordonez2016deep} have introduced a hybrid network called DeepConvLSTM, comprised of CNNs and Long Short-Term Memory (LSTM) units, which could simultaneously extract local and global feature representations from multimodal sensor signals for HAR. Sharma et al. \cite{sharma2022transformer} have developed a Transformer-based architecture for modeling long-range dependence from sensor data, demonstrating a strong feature fusion ability for activity classification. Despite their outstanding success in the ubiquitous HAR scenario, most existing deep learning models typically assume that the training and test sets follow the same distribution, and neglect the fact that the sensor data collected from various subjects may have diverse distributions because of their unique biological characteristics such as body shapes, genders, and behavior habits, hence rendering suboptimal activity recognition performance under non-IID setting \cite{huang2022channel}.

\subsection{Domain Adaption and Generalization}
To solve such distribution-shift problem, domain adaptation (DA) has emerged as a promising solution to bridge domain gaps, which attempts to leverage the source domain to improve activity recognition performance on unlabeled or unannotated target domain data. For instance, Wang et al. \cite{wang2018stratified} have introduced a stratified transfer learning algorithm called STL to learn class-wise feature transformations for cross-domain activity recognition tasks. Lu et al. \cite{lu2021cross} have utilized a substructure-level optimal transport technique to improve cross-domain activity recognition. In addition, Qin et al. \cite{qin2019cross} have presented a novel approach named ASTTL, which employs an adaptive spatial-temporal transfer learning for cross-dataset activity recognition. However, DA remains a major limitation since it often requires direct access to the target domain during model training, which renders it infeasible or impractical in real-world HAR deployment. To overcome this drawback, domain generalization (DG) has provided an alternative solution, which aims to learn transferable and common knowledge from one or multiple different but related source domains, which can generalize well on new never-seen-before target domains without requiring direct access to them. Existing domain generalization (DG) works can be broadly divided into three categories: Representation learning \cite{lu2024diversify,qian2021latent}, Data augmentation \cite{zhang2018mixup,lu2024fixed,du2021adarnn}, and Learning strategies \cite{huang2020self,sagawadistributionally}. Though many effective DG algorithms have been successfully applied in computer vision (CV) area, they could not be well compatible with time series data \cite{wang2022generalizing,zhou2022domain}. So far, there has existed very limited DG-based research attention for human activity recognition utilizing wearable sensor data. For instance, Du et al. \cite{du2021adarnn} have introduced an Adaptive RNN named AdaRNN to tackle the distribution-shift problem via building an adaptive model, that is able to generalize well on unseen time series test data. Qian et al. \cite{qian2021latent} have proposed a new approach called Generalizable Independent Latent Excitation (GILE) to automatically disentangle domain-invariant and domain-specific features, which considerably improves the generalization capability of the model for cross-person activity recognition. However, their use of generative models for feature disentanglement inevitably incurs additional inference burdens. Lu et al. \cite{lu2024fixed} have developed a simple yet effective domain-invariant Feature Mixup with Enhanced Discrimination algorithm called FIXED to enhance domain-invariance for DG-based activity recognition. To the best of our knowledge, DIVERSIFY is the latest work, that has attempted to employ domain-invariant representation learning techniques for time series data out-of-distribution detection and generalization \cite{lu2024diversify}. Overall, our work is clearly distinct from all the methods mentioned above, which formulates such DG-based HAR problem from a new viewpoint of causal inference, by excavating the underlying causal mechanism to disentangle causal and non-causal representations contained in raw sensor signals, indicating stronger generalization ability. Unlike existing techniques, this paper is the first to consider the disentanglement between causal and non-causal features, emphasizing the effectiveness of causal features in this context of HAR.

\subsection{Causal Representation Learning in DG-Based HAR}
Disentangled representation learning has recently emerged as a new metrology for domain generalization, which attempts to learning representations where informative and distinct factors of data variation may be effectively separated. It has increasingly garnered research attention in computer vision community \cite{liu2021towards}. To our knowledge, the most prevalent solutions for feature disentanglement are mainly based on Variational Autoencoders (VAE) \cite{higgins2017beta,kim2018disentangling}, which can be performed in an entirely unsupervised manner. Particularly, a new research line of disentangled representation learning named causal representation learning has arisen. For instance, Yang et al. \cite{yang2021causalvae} have introduced CausalVAE, which combines the linear Structural Causal Model (SCM) with VAE to learn the latent representation with causal structure. Zhang et al. \cite{zhang2022towards} have presented a primal-dual method utilizing joint representation disentanglement for domain generalization, which demonstrate a great potential of disentanglement in enhancing generalization performance. Lv et al. \cite{lv2022causality} have suggested a general structural causal model to extract causal factors from input data, which can help to capture invariant causal mechanism. Wang et al. \cite{wang2022out} have introduced a causal invariant transformation, which can exploit the invariant properties to learn a generalizable model, instead of explicitly recovering the causal features. Jiang and Veitch \cite{jiang2022invariant} have proposed to learn invariant and transferable representations, based on the shared causal structure of domains. Despite being effective, most existing causality representation learning literatures mainly concentrate on computer vision tasks \cite{liu2021towards}. So far, there have existed only a few works that adopt casualty learning for human activity recognition. For example, Kim et al. \cite{kim2023causality} have proposed a causality-aware pattern mining strategy for group activity recognition. Hwang et al. \cite{hwang2021deep} have presented a deep learning framework for human activity recognition based on causal feature extraction. Lai et al. \cite{lai2021capturing} have introduced a causal network to capture the effect of biases on activity recognition performance and show how knowledge distillation could be adopted to mitigate such bias influence. However, they have paid less attention to domain generalization in the context of sensor-based activity recognition. Unlike existing methods, this paper is the first to consider the disentanglement representation learning to separate causal and non-causal features, emphasizing the effectiveness of causal representation learning in generalizable sensor-based activity recognition scenario.

\section{METHOD}

\subsection{Preliminaries}
To tackle activity recognition problem under such domain-shift setting, we typically denote the sensor data space by $\mathcal{X}$ and the activity label space by $\mathcal{Y}$. Given that there is a total of $S$ source domains denoted as $\mathcal{D}_{tr} = \{D_1, D_2, \cdots, D_S\}$, for the $i$-th source domain $\mathcal{D}_i$, the joint distribution may be represented as $P^{i}(x, y)$, where $x \in \mathcal{X}$ refers to the sensor data inputs obtained through a popular sliding window technique \footnote{Due to its implementational simplicity, a popuar sliding window strategy is used to divide continous time series data into fixed-size windows as sensor inputs here \cite{ordonez2016deep}.}, and $y \in \mathcal{Y} = \{1, 2, \cdots, K\}$ denotes that there are $K$ categories in the activity label space. The goal of the DG-based HAR task is to learn a model $H$ from the source domains $\mathcal{D}_{tr}$ to recognize activity categories in the unseen target domain $\mathcal{D}_{te}$. It is important to note that the target domain $\mathcal{D}_{te}$ can only be accessed during inference. We aim to train the model $H$ on the source domains $\mathcal{D}_{tr}$ to minimize the average prediction error $\epsilon_t$ on the target domain $\mathcal{D}_{te}$:
\begin{equation}
	\begin{aligned}
		\label{eq_1}
		\epsilon_t = \mathbb{E}_{(\boldsymbol{x}, y) \sim {P}^{te}(\boldsymbol{x}, y)}\mathcal{L} (H(\mathbf{x}), y).
	\end{aligned}
\end{equation}

\begin{figure*}[!t]
	\centering
	\includegraphics[width=1.0\linewidth]{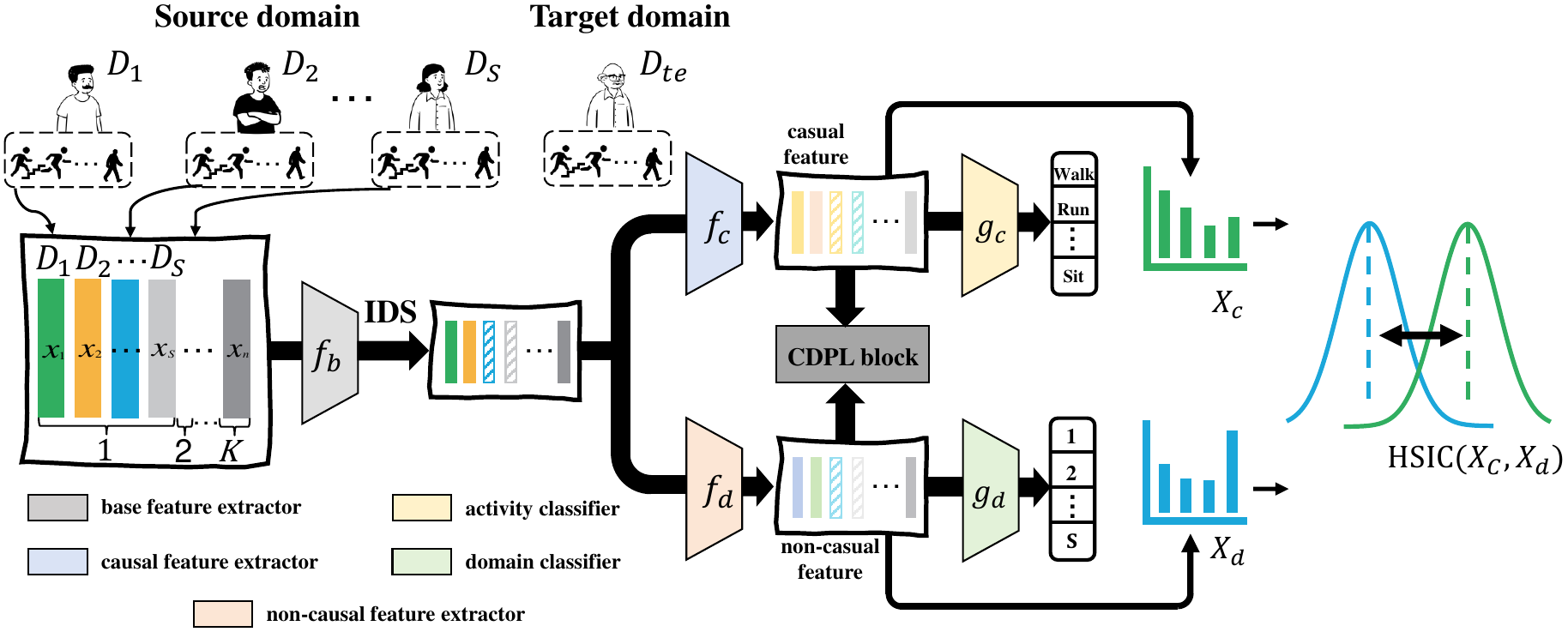}  
	\caption{The Early-Forking Two-Branch framework is structured as follows. First, sensor data for activity recognition from the source domain are processed through a base feature extractor $f_b$ to obtain foundational features, which are then augmented using the IDS strategy. Causal features $X_c$ are extracted through the upper branch, while non-causal features $X_d$ are extracted through the lower branch. HSIC is applied to reduce dependency between the two branches, and a CDPL block is incorporated to prevent feature representation collapse. The upper branch performs activity recognition classification, and the lower branch conducts domain-invariance learning. During inference, only the upper branch is utilized for activity classification.}
	\label{fig_network}
\end{figure*}

\subsection{Overall Framework} \label{overall_framework}
Based on the above formulation, we present an overview of our causal relationship-inspired method for activity recognition based on Domain Generalization (DG). Our primary hypothesis is that, despite the entanglement of causal and non-causal features, we can realize their disentanglement by aggregating activity data from multiple source domains with diverse distributions. This intuition motivates us to learn causal features that generalize well in the target domain. This hypothesis is quite reasonable: different individuals with diverse biological traits and behavioral habits may perform the same activity but yield distinct sensor readings, indicating an underlying causal relationship among them. Consequently, we can leverage causal relationship-inspired methods to train our Human Activity Recognition (HAR) model to learn domain-invariant representations of causal features. The core idea is to capture transferable knowledge of causal features from multiple source domains to enhance generalization capability. Our overall framework is illustrated in \figurename~\ref{fig_network}. Concretely, our structural design follows two main guidance principles:

First, it should be assured there exists no association between causal and non-causal features from the same activity category, i.e., $X_c \perp \!\!\! \perp X_d$. Actually, in a classical two-branch architecture, a popular design is to apply a shared base feature extractor \cite{atzmon2020causal,lv2022causality,chen2024causal}, which is then attached with two lightweight classification heads. However, such design would be detrimental to generalizable activity classification performance, because it may cause the causal feature extractor to heavily rely on the non-causal feature extractor. As an alternative, an early-forking two-branch framework is suggested, where both of branches only share the initial few blocks, while diverging thereafter along this two-branch backbone structure, as illustrated in \figurename~\ref{fig_network}. In such a case, though the low-level features of causal and non-causal factors may be entangled, their high-level features would be divergent. On this basis, the Hilbert-Schmidt Independence Criterion (HSIC) is employed as a metric of feature independence to disentangle the causal and non-causal features\cite{gretton2007kernel,greenfeld2020robust}.

Second, the causal features from the same activity category should remain invariant across different domains, i.e., $X_c \perp \!\!\! \perp X_d|C$. As mentioned in prior reference \cite{atzmon2020causal}, since the training samples themselves are usually labeled in a compositional manner, directly disentangling their “elementary” components from samples is typically an ill-posed problem. For example, in such activity recognition scenario, there are two persons, one with a height of 190 cm and the other with a height of 160 cm, who perform “Running” exercise. Through deep models that learn from sensor data, the resulting representations of those components would be inherently entangled. Thus, it is difficult to directly tell which features indicate the subject height and which features indicate the activity class, i.e., “Running”. How to identify ideal casual feature $X_c$ from only limited observed sensor-label pairs $(x, y)$, which is known to be an ill-posed problem. Because of such ill-posed property of this problem, it might be still insufficient to find the optimal causal features through merely enforcing the aforementioned feature independence constraint alone (i.e., $X_c \perp \!\!\! \perp X_d$). Enforcing more disentanglements of the learned representations would be of benefit (i.e., $X_c \perp \!\!\! \perp D|C$). To further strength feature disentanglement, we propose to impose more constraint terms on the causal features. To implement the second guidance idea (i.e., $X_c \perp \!\!\! \perp D|C$), we treat each sensor input as a distinct activity category and perform data augmentation to generate samples from different domains that belong to the same category. To this end, a new inhomogeneous domain sampling (IDS) technique is introduced to expand domain styles. Different form prior arts \cite{zhoudomain,li2022uncertainty} that synthesize new domains through the addition of noise or mixing techniques, our IDS is able to directly generate diverse domains with different styles via simply perturbing the feature maps with controlled intensity. In such a way, comparing to original sensor input, the generated data would contain different non-causal factors $X_d$ but the same causal-factors $X_c$, hence forcing the causal features to be domain-invariant.

Moreover, following the previous reference \cite{chen2021exploring,kim2021selfreg}, it is important to note that here a category-aware domain perturbation layer named CDPL (i.e., a MLP projection layer) is employed to prevent representation collapse along the two branches. Specifically, borrowing an idea of Siamese networks \cite{chen2021exploring}, we place an additional CDPL projection layer at the end of our two-branch framework, which includes two fully-connected layer, among which both batch normalization (BN) operation and ReLU activation are inserted. It is mainly responsible for avoiding representation collapse, when one sensor sample is simultaneously processed by the two branches to form two augmented views (i.e., casual and non-casual) without using any negative-pair samples. We will provide experimental evidence to support our structural design in later sections.

\subsection{The Two-Branch Structure with Early-Forking}
As previously introduced, we employ a two-branch framework with early-forking after a shared base feature extractor $f_b$. The upper branch is designated for extracting causal features, while the lower branch focuses on non-causal features. In the upper branch, the causal feature extractor is defined as a composite function $F_c=f_c (f_b (x))$, where $f_c (\cdot)$  is in charge of extracting high-level casual features. Similarly, in the lower branch, the non-causal feature extractor is defined as $F_d=f_d (f_b (x))$, where $f_d (\cdot)$  is in charge of extracting high-level non-casual features. Since deep learning models usually learn common features in their lower layers, such early-forking design is reasonable, which only shares the first lower layers (i.e., $f_b (x)$) and diverges thereafter to perform two-branch structure (i.e., $f_c (f_b(x))$ and $f_d (f_b(x)))$. This framework enables implicit independence regularization on $X_c$ and $X_d$ to disentangle causal and non-causal features, as depicted in \figurename~\ref{fig_mt}. Since $X_c$ and $X_d$ are unobservable, we follow a common practice \cite{christiansen2021causal,atzmon2020causal,chen2024causal} by learning them from $F_c$ and $F_d: \mathcal{X} \rightarrow \mathcal{Z}$, where $\mathcal{Z}$ represents the feature space. We will utilize the Hilbert-Schmidt Information Criterion (HSIC) \cite{atzmon2020causal,greenfeld2020robust} as an independence measure to regularize the statistical dependency between the causal features extracted by $F_c$ and the non-causal features extracted by $F_d$, effectively facilitating the disentanglement of these features. During training, both branches are utilized; however, only the upper branch is employed during inference. In subsequent sections, we will provide an ablation study to validate the effectiveness of HSIC. Our final activity classifier $g_c$, and domain classifier $g_d$ are constructed using simple fully connected layers, which are employed to predict activity and domain labels, respectively. Here domain labels respectively refer to different subject IDs, sensor device locations, and dataset IDs in cross-person \cite{qian2021latent,lu2024diversify,lu2024fixed}, cross-position \cite{lu2024diversify}, and cross-dataset \cite{lu2024diversify} scenarios, which can often be directly accessible via metadata or experimental setups. Specifically, for the cross-person scenarios in public HAR benchmark datasets like DSADS \cite{altun2010comparative} and PAMAP2 \cite{reiss2012introducing}, the domain labels are inherently embedded in subject IDs, with each person having a unique ID. In the cross-position setting, wearable sensors are placed at five distinct body locations, each of which corresponds to a unique sensor location ID. Similar case also happens in the cross-dataset setting, where each dataset is assigned with a unique dataset ID. Since both the user identities and sensor deployments (i.e., domain labels) of these recruited subjects must be accurately registered to ensure reliable sensor data label collection, the domain labels are often known. Below, we will discuss how our causal relationship-inspired method addresses such DG-based HAR problem.

\begin{figure}[!t]
	\centering
	\subfigure[]{
		\includegraphics[width=0.42\linewidth]{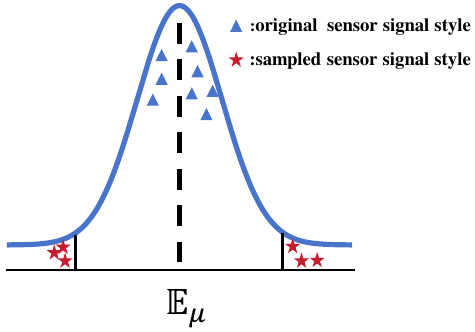}
		\label{fig_ds}
	}
	\subfigure[]{
		\includegraphics[width=0.36\linewidth]{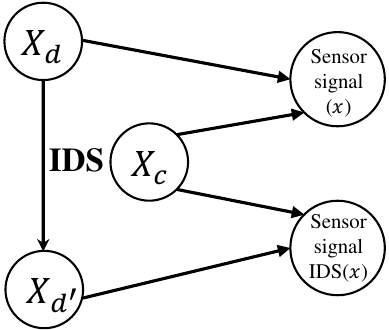}
		\label{fig_aug}
	}
	\vspace{-.2in}
	\caption{\textbf{(a)} Inhomogeneous sampling of sensor signal styles. \textbf{(b)} To augment the non-causal domain information, an inhomogeneous sampling strategy is performed to produce new sensor sample IDS$(x)$ from original sensor sample $x$, which makes the resulting non-causal feature  $X_{d'}$ differ from $X_d$ while preserving the causal feature $X_c$ to be invariant across domains.}
	\label{fig_ids}
\end{figure}

\subsection{Hilbert-Schmidt Information Criterion}
The two-branch framework is employed to decouple causal and non-causal features, corresponding to $X_c \perp \!\!\! \perp X_d$ through d-separation using a structural causal model as shown in \figurename~\ref{fig_mt}. Since both $X_c$ and $X_d$ cannot be explicitly formulated, we learn them separately through $F_c$ and $F_d$: $\mathcal X \to \mathcal Z$. In fact, Gretton et al. have shown that Hilbert-Schmidt Information Criterion (HSIC) is, indeed, an ideal indicator for the independence of random variables \cite{gretton2005measuring}. Following previous literature \cite{atzmon2020causal,chen2024causal,greenfeld2020robust}, we then employ the Hilbert-Schmidt Information Criterion (HSIC) as a regularization term to disentangle the causal features $X_c$ from the non-causal features $X_d$. HSIC has been a popular non-parametric approach for estimating the statistical dependence between samples of two random variables, which is mainly based on an implicit embedding into a universal reproducing kernel Hilbert space (RKHS). In the infinite-sample limit, the HSIC between two random variables will be 0 if and only if both of them are independent. HSIC can also serve as a simple finite-sample estimator, which is differentiable with regards to input variables. Due to its non-parametric advantage, HSIC does not need to train an additional network, which is very easy to optimize. The suggested independence criterion is mainly based on the eigenspectrum of covariance operators in reproducing kernel Hilbert spaces (RKHSs), which consists of an empirical estimate of the Hilbert-Schmidt norm of the cross-covariance operator. The loss minimization function $F_c(x)$ for disentanglement can be expressed as:
\begin{equation}
	\begin{aligned}
		\label{eq_2}
		arg \min_{F_c(x)}\mathrm{HSIC}(F_c(x),F_d(x))=arg\min_{F_c{(x)}}{1\over{(B-1)^2}}\mathbf{trace(KHLH),}
	\end{aligned}
\end{equation}
where $B$ is the batch size; Referring to this reference \cite{gretton2005measuring}, we adopt the trace of KHLH to achieve disentanglement, where $\mathbf{trace({ \cdot })}$ computes the trace of the given matrix. We have chosen a linear kernel instead of a Gaussian kernel, as the linear kernel does not involve complex parameter selection issue \cite{atzmon2020causal,chen2024causal,greenfeld2020robust}. Here, $\mathbf{K}= F_c(x)' F_c(x)'^{\text{T}}$, $\mathbf{H}=\mathbf{I}-{1\over B}$ is the centering matrix, and $\mathbf{L}=F_d(x)' F_d(x)'^{\text{T}}$. With regards to $F_c (x)$, we formulate $F_c (x)'$ as $F_c (x)'={F_c (x) \over ||F_c (x)||}$, and similar for $F_d (x)'$. Though one might argue that there exist other independence constraints such as orthogonal or correlation minimization, we show that HSIC is more effective compared to other alternatives in later Section \ref{sec:ablation}.

\subsection{Inhomogeneous Domain Sampling}
As previously mentioned, since using HSIC alone is insufficient to fully disentangle causal and non-causal features, imposing additional constraint terms are necessary \cite{greenfeld2020robust,gretton2007kernel}. First, we argue that the branch in charge of extracting causal features should not depend on the branch in charge of extracting non-causal features. The causal feature extractor and the non-causal feature extractor only partially share the base feature extraction layers responsible for capturing low-level features, after which the two extraction branches would be divergent. Second, based on the assumption that the causal features of activity samples belonging to the same category across different domains remain consistent, additional constraint terms should be imposed. These constraints would not only enforce similarity between activity samples of the same class across different domains, but also ensure that the causal features of the original samples and their augmented versions are identical.
To maintain consistent causal features across different domains for the same sample, we introduce an inhomogeneous domain sampling (IDS) strategy. This strategy allows the mixing of domain information from two different samples to create a new domain. For example, given a set of feature maps with a size of $B \times C \times 1 \times W$, where $B$, $C$, 1, and $W$ represent the batch size, channel number, height, and width of the feature maps respectively, the height of temporal data is equal to 1 after preprocessing through a sliding window. Domain information is represented by the mean $\mu \in \mathbb{R}^{B\times C}$ and variance $\sigma^2 \in \mathbb{R}^{B \times C }$ of activation values across all spatial positions and all activity samples in current one batch \cite{zhoudomain,wang2024optimization}.

\begin{algorithm}[!t]
	\renewcommand{\algorithmicrequire}{\textbf{Input:}}
	\renewcommand{\algorithmicensure}{\textbf{Output:}}
	\caption{Causality-inspired representation learning for DG-based HAR}
	\label{alg:1}
	\begin{algorithmic}[1]
		\REQUIRE The training domain $\mathcal{D}_{tr}$, and hyper-parameters $\alpha,\beta$;
		\ENSURE The trained model $H$;
		\STATE Randomly initialize the model parameters;
		\WHILE {not converge}{
			\STATE Sample a mini-batch $\mathcal{B} \gets\ \{\mathcal{B}^1,\mathcal{B}^2...,\mathcal{B}^i\}$ from $i$ domains;
			\STATE Extract the base features $f_b(x)$ by the base features extractor;
			\STATE Inhomogeneous Domain Sampling according to Equation (\ref{eq_5});
			\STATE Extract the causal features $f_c(f_b(x))$ by causal features extractor;
			\STATE Extract the non-causal features $f_d(f_b(x))$ by non-causal features extractor;
			\STATE Calculate the independence loss $\mathcal{L}_\text{ind}$ according to Equation (\ref{eq_7});
			\STATE Calculate the consistency constraint loss $\mathcal{L}_\text{con}$  according to Equation (\ref{eq_6});
			\STATE Calculate the cross-entropy loss $\mathcal{L}_\text{cls}$ according to Equation (\ref{eq_8});
			\STATE Calculate the total loss $\mathcal{L}$ according to Equation (\ref{eq_9});
			\STATE Update the model parameter using Adam optimizer;
		}
		\ENDWHILE\\
	\end{algorithmic}
\end{algorithm}

In our design as depicted in \figurename~\ref{fig_network}, we hope that our domain sampling strategy may modify sensor samples in such a way, which is able to preserve the casual semantics (Category) while adjusting the non-causal domain information for a given sensor sample. When dealing with similar domain types in the current batch, directly modifying $\mu$ and $\sigma^2$ by mixing with other domain styles might lead to relatively uniform features, producing too homogeneous non-causal features $\mathbf{X}_d$. IDS mitigates this by introducing controlled perturbations to domain styles. Specifically, the mean and variance of activations for each sensor data sample, denoted as $\mu_b$ and $\sigma_b^2$, should be first computed. Then, the probabilistic models for $\mu_b$ and $\sigma_b^2$ are constructed by assuming they follow a Gaussian distribution, with batch-level model parameters estimated. New $\mu$ and $\sigma^2$ values are sampled from the Gaussian distribution to represent a new domain style. For $\mu$, this process can be formulated as:
\begin{equation}
	\label{eq_3}
	\mu \sim \mathcal{N}\left(\mathbb{E}_\mu, \Sigma_\mu\right) \text {, s.t. }\left\{\begin{array}{l}\mathbb{E}_\mu=\frac{1}{B} \sum_{b=1}^B \mu_b, \\\Sigma_\mu=\frac{1}{B} \sum_{b=1}^B\left(\mu_b-\mathbb{E}_\mu\right)\left(\mu_b-\mathbb{E}_\mu\right)^{\mathrm{T}} .\end{array}\right.
\end{equation}
To differentiate from uniform domain styles, we sample $\mu$ values with a density less than $\epsilon$, where those accepted samples $\overline \mu$ should satisfy:
\begin{equation}
	\label{eq_4}
	\frac{1}{(2 \pi)^{\frac{B \times C}{2}}\left|\Sigma_\mu\right|^{\frac{1}{2}}} \exp \left(-\frac{1}{2}\left(\overline{\mu}-\mathbb{E}_\mu\right)^{\mathrm{T}} \Sigma_\mu^{-1}\left(\overline{\mu}-\mathbb{E}_\mu\right)\right)<\epsilon.
\end{equation}
Here, $\epsilon$ is chosen to be sufficiently small so as to ensure that the sampled $\overline{\mu}$ differs from the original $\mu$. The same process is also applied to sample $\sigma^2$. Once the new domain style is generated, the original domain style is replaced with the new one, and the final formula is written as:
\begin{equation}
	\label{eq_5}
	\text{IDS} {(x)} = \overline{\sigma} \frac{f_b(x) - \mu}{\sigma} + \overline{\mu}.
\end{equation}
Additional consistency constraint term should then be imposed, which forces the causal features of the original sample and its corresponding sampled version to remain equal. This process can be expressed as:
\begin{equation}
	\label{eq_6}
	\begin{split}
		\mathcal{L}{_\text{con}}=\Vert F_{c}{(x)- \text{CDPL}(f_c(\text{IDS}(x)))}\Vert_1-  \Vert \text{CDPL}(F_c(x))-f_{c}({\text{IDS}(x)})\Vert_1- \\ \text {min}(([\Vert F_d(x) - \text{CDPL}(f_d(\text{IDS}(x)))\Vert_1 - m],0) +  \text{min}([\Vert \text{CDPL}(F_d(x)) - f_d(\text{IDS}(x)) \Vert_1 - m],0)),
	\end{split}
\end{equation}
where $\Vert \cdot \Vert_1$ denotes the $\mathcal{L}_{1}$-norm, and $m$ is the maximum $\mathcal{L}_{1}$ distance computed from the current batch. In line with previous work \cite{atzmon2020causal}, it is worth noting that here a category-aware domain perturbation layer (CDPL) \cite{zbontar2021barlow,chen2021exploring} is employed to prevent representation collapse along the two branches. Distinct from prior works \cite{zhoudomain,li2022uncertainty} that concentrate on enforcing normal perturbations to domain style or borrowing them from other domains, our domain sampling strategy could better control how the generated domain styles deviate from the existing ones due to its probabilistic modeling of $\mu_b$ and $\sigma^2_b$, hence avoiding producing too homogeneous domains in such data augmentation process.

\begin{algorithm}[!t]
	\renewcommand{\algorithmicrequire}{\textbf{Input:}}
	\renewcommand{\algorithmicensure}{\textbf{Output:}}
	\caption{Inhomogeneous Domain Sampling}
	\label{alg:2}
	\begin{algorithmic}[1]
		\REQUIRE The feature map $z\in \mathbb{R}^{B\times C\times 1\times W}$ is obtained after passing through the base feature extractor $f_b(x)$;
		
		//\textcolor{gray}{As an example, in the cross-person task of the USC-HAD dataset, the shape of $z$ is (96, 32, 1, 58), with a batch size of 32. The value 96 represents the use of data from three source domains for training;}
		\ENSURE The $\overline{z} \in \mathbb{R}^{B\times C\times 1\times W}$ obtained through Inhomogeneous Domain Sampling;
		\STATE  The mean and variance for a mini-batch are computed along dimensions 2 and 3;
		
		$\mu \gets\ \frac{1}{W}\sum\limits_{w=1}^{W} z_{b,c,1,w}$, $\sigma^2 \gets\ \frac{1}{W}\sum\limits_{w=1}^{W} (z_{b,c,1,w} - \mu)^2$; 
		
		//\textcolor{gray}{In the USC-HAD dataset, both $\mu $ and $ \sigma^2 $ have a shape of (96, 32, 1, 1);}
		\STATE The $x$ is normalized using the mean $\mu$ and variance $\sigma^2$;

		\STATE Construct a probabilistic model for $\mu$.
		
		$\mu\sim\mathcal{N}(\mathbb{E}_\mu,\Sigma_\mu),\mathrm{s.t.}\left\{
		\begin{array}
			{l}\mathbb{E}_\mu \gets\ \frac{1}{B}\sum_{b=1}^B\mu_b, \\
			\Sigma_\mu  \gets\ \frac{1}{B}\sum_{b=1}^B(\mu_b-\mathbb{E}_\mu)(\mu_b-\mathbb{E}_\mu)^\mathrm{T}.
		\end{array}\right.$
		
		\STATE We sample $\overline{\mu}$ values with a density less than $\epsilon$;
		
		$\frac{1}{(2 \pi)^{\frac{B \times C}{2}}\left|\Sigma_\mu\right|^{\frac{1}{2}}} \exp \left(-\frac{1}{2}\left(\overline{\mu}-\mathbb{E}_\mu\right)^{\mathrm{T}} \Sigma_\mu^{-1}\left(\overline{\mu}-\mathbb{E}_\mu\right)\right)<\epsilon.$
		
		//\textcolor{gray}{In the USC-HAD dataset, we set $\epsilon$ to be sufficiently small, with a value of 0.0001;}
		\STATE Repeat the above two steps to construct a probabilistic model for $\sigma$, and sample $ \overline{\sigma}$;
		\STATE Scale and shift the normalized $z$ using the sampled $\overline{\mu}$ and $\overline{\sigma}$;
		
		$\overline z \gets\ \overline{\sigma} \frac{z - \mu}{\sigma} + \overline{\mu}.$
		
	\end{algorithmic}
\end{algorithm}

\subsection{Learning Objective}
After performing inhomogeneous domain sampling (IDS), the samples would be disentangled. The disentanglement loss function can be expressed as:
\begin{equation}
	\label{eq_7}
	\mathcal{L}_\text{ind}=\sum_{\{{x_1,x_2\}\in{\{x,\text {IDS}(x)\}}}}\text{HSIC}(F_c(x_1),F_d(x_2)).
\end{equation}
The standard classification loss $\mathcal{L}_\text{cls}$ is defined as:
\begin{equation}
	\label{eq_8}
	\mathcal{L}_\text{cls}=\sum \mathrm{CE}(g_c(F_c(x_1)),y)+\mathrm{CE}(g_d(F_d(x_2)),d),
\end{equation}
where $\mathrm{CE}$ represents the cross-entropy loss, $y$ denotes the activity label, and $d$ denotes the domain label.The overall learning objective can be formulated as:
\begin{equation}
	\label{eq_9}
	\mathcal{L} = \mathcal{L}_\text{cls} + \alpha \mathcal{L}_\text{ind} + \beta \mathcal{L}_\text{con},
\end{equation}
where $\alpha$ and $\beta$ are balancing factors that control the trade-off between the two loss terms defined in Eq. (\ref{eq_6}) and Eq. (\ref{eq_7}). In summary, the proposed method utilizes a two-branch framework with early-forking during the training phase. During inference, the branch responsible for domain classification based on the non-casual features is removed, leaving only the other branch for activity recognition. Our design incurs no additional computation or memory cost during inference, ensuring that the actual activity recognition time is not increased. A detailed evaluation of the activity inference latency will be provided in the subsequent section \ref{infer}. The complete learning procedure is presented in Algorithm \ref{alg:1}. To make such inhomogeneous domain sampling (IDS) strategy more clear, we provide an exemplar pseudo-code, as illustrated in Algorithm \ref{alg:2}.

\section{Experiments}

\subsection{Datasets}
This section provides a detailed description of four publicly available Human Activity Recognition (HAR) benchmark datasets. The DSADS dataset \cite{altun2010comparative} utilizes multiple sensors to collect information from five positions on different body parts, encompassing 19 activities performed by a total of 8 subjects, aged between 20 and 30 years. The USC-HAD dataset  \cite{zhang2012usc} employs a sensor attached to the right anterior hip to record 12 kinds of activities, involving 14 subjects (7 males and 7 females) aged between 21 and 49 years. The PAMAP dataset \cite{reiss2012introducing} also utilizes multiple sensors to gather information from three body locations, covering data from 12 types of activities performed by 10 subjects, all aged between 20 and 30 years. The UCI-HAR dataset \cite{anguita2013public} contains sensor information collected from a smartphone worn at the waist, encompassing 6 activities of daily living, with all subjects roughly aged between 19 and 48 years. Additional information can be found in \tablename~\ref{tab_dataset}.
\begin{table}[!t]
	\small
	\caption{Data preprocessing and experimental setting.}
	\renewcommand{\arraystretch}{1.0} 
		\begin{tabular}{@{}ccccc@{}}
			\toprule
			Attribute  & DSADS & USC-HAD & PAMAP2 &UCI-HAR \\ \midrule
			Activity Classes & 19 & 12 & 12 & 6 \\ \midrule
			Subjects & 8 & 14 & 9 & 30\\ \midrule
			Sample Rate(Hz) & 25 & 100 & 100 & 50 \\ \midrule
			Activity Classes &\begin{tabular}[c]{@{}c@{}}Sitting\\Standing\\Lying on back\\etc.\\\end{tabular} &
			\begin{tabular}[c]{@{}c@{}}Jumping\\Sitting\\Standing\\etc.\\\end{tabular} &
			\begin{tabular}[c]{@{}c@{}}Lying\\Sitting\\Standing\\etc.\\\end{tabular} &
			\begin{tabular}[c]{@{}c@{}}Walking\\Sitting\\Standing\\etc.\\\end{tabular}  \\  \midrule
			Sensor Modalities  & \begin{tabular}[c]{@{}c@{}}Accelerometer\\Gyroscope\\magnetometer\\\end{tabular} & \begin{tabular}[c]{@{}c@{}}Accelerometer\\Gyroscope\\magnetometer\\\end{tabular}  & \begin{tabular}[c]{@{}c@{}}Accelerometer\\Gyroscope\\magnetometer \\\end{tabular} &
			\begin{tabular}[c]{@{}c@{}}Accelerometer\\Gyroscope\\- \\\end{tabular}\\ \midrule
			Sensor Location &\begin{tabular}[c]{@{}c@{}}Torso\\Right arm\\Right leg\\etc.\\\end{tabular} &
			\begin{tabular}[c]{@{}c@{}}Front right hip\\-\\-\\-\\\end{tabular} &
			\begin{tabular}[c]{@{}c@{}}Wrist on the dominant arm\\ Dominant side's ankle\\Chest\\-\\\end{tabular} &
			\begin{tabular}[c]{@{}c@{}}Waist\\-\\-\\-\\\end{tabular}  \\
			\bottomrule
		\end{tabular}
	\label{tab_dataset}
\end{table}

\begin{table}[]
	\small
	\caption{Detailed cross-domain setting in sensor-based HAR.}
	\resizebox{0.95\textwidth}{!}{
	\begin{tabular}{cccccc}
		\toprule
		\multicolumn{6}{c}{Cross-person setting}                                                                                                  \\ \midrule
		Dataset & Window size & Overlap rate & Domain & Domain ID                         & Domain sample                                 \\ \midrule
		DSADS   & 45×125      & 50\%         & 4      & [0, 1], [2, 3], [4, 5], [6, 7]                 & (28,500)×4                                 \\ \midrule
		USC-HAD & 6×200       & 50\%         & 4      & [1, 11, 2, 0], [6, 3, 9, 5], [7, 13, 8, 10], [4, 12] & (1,401,400;1,478,000;1,522,800;1,038,800) \\ \midrule
		PAMAP2  & 27×200      & 50\%         & 4      & [3, 2, 8], [1, 5], [0, 7], [4, 6]               & (592,600;622,200;620,000;623,400)            \\ \midrule
		\multicolumn{6}{c}{Cross-position setting}                                                                                                \\ \midrule
		Dataset & Window size & Overlap rate & Domain & Domain ID                         & Domain sample                              Total     \\ \midrule
		DSADS   & 9×125       & 50\%         & 5      & Five distinct domains based on different sensor locations.                                 & (1,140,000)×5                             \\ \midrule
		\multicolumn{6}{c}{Cross-dataset setting}                                                                                                 \\ \midrule
		Dataset & Window size & Overlap rate & Domain & Domain ID                         & Domain sample                              Total     \\ \midrule
		-       & 6×50        & 50\%         & 4      & Each dataset is treated as a domain.                                 & (672,000;810,550;514,950;470,850)         \\ \bottomrule
	\end{tabular}
}
\end{table}

\subsection{Cross-Domain Settings}
In this section, the four distinct cross-domain settings utilized in our study are introduced in details.
\begin{enumerate}
\item \textbf{Cross-person setting \footnote{Since the baselines for UCI-HAR are already good enough, we do not run cross-person and one-to-the-other-person experiments on it.} \cite{lu2024diversify,qian2021latent}:} The DSADS, USC-HAD, and PAMAP2 datasets were selected for cross-person experiments. For every dataset, we split the whole data into four domains according to the persons, with each domain having a similar amount of data. For instance, there is a total of 8 persons in the DSADS dataset, which may be divided into four groups or domains: [0, 1], [2, 3], [4, 5], [6, 7], where different numbers represent different persons.
\item\textbf{Cross-position setting \cite{lu2024diversify}:}  For cross-position experiments, the DSADS dataset, which features the most extensive range of sensor placement locations, was utilized. The dataset was grouped into five distinct domains based on different sensor locations: Torso, Right arm, Left arm, Right leg, Left leg.
\item\textbf{Cross-dataset setting \cite{lu2024diversify}:} To construct the cross-dataset scenario, we merged four datasets including DSADS, USC-HAD, PAMAP2, and UCI-HAR, where each dataset could be treated as one domain, resulting in a total of four domains. Intuitively, due to being more diverse (e.g., different subjects groups, device deployment, device type, sampling rate, etc.), such cross-dataset setting should be more challenging than cross-position setting, which means that not only the sensor positions and types are different, but also the persons are different. To make such cross-dataset setting practically feasible, following the common practice in previous literature \cite{lu2024diversify}, we selected six common activities, that are present across all four datasets. Moreover, these activities originating from gyroscope and accelerometer data were collected at similar or identical positions in each dataset. The high-frequency data from the four datasets were downsampled to 25 Hz, ensuring a consistent sampling frequency across all datasets. Under this premise, the cross-position setting would be more challenging than such constructed cross-dataset setting.  
\item\textbf{One-person-to-another setting \cite{lu2024diversify}:} The DSADS, USC-HAD, and PAMAP2 datasets were employed to create such one-person-to-another setting. Following previous literatures \cite{lu2024diversify}, we selected the subjects with IDs 0–7 from the DSADS, USC-HAD, and PAMAP2 datasets to form four subject pairs: (0←1), (2←3), (4←5), and (6←7). In each pair, the former subject serves as the test set, while the later subject serves as the training set. In fact, the aforementioned cross-person setting mainly attempt to generalizing from multiple source domains to one target domain. In contrast to the cross-person setting, the one-person-to-another setting concentrates more on single-source cross-person generalization, which generalizes from one subject to another. Overall, such one-person-to-another setting is more challenging, since it contains fewer sensor samples.
\end{enumerate}

\subsection{Implementation Details}
As previously mentioned in \figurename~\ref{fig_network}, our backbone architecture comprises a base feature extractor, a causal feature extractor, a non-causal feature extractor, an activity classifier, and a domain classifier. Following previous references \cite{wang2019deep,lu2024diversify,lu2024fixed}, the base feature extractor consists of a convolutional layer followed by a max pooling layer. Both the causal and non-causal feature extractors are structured similarly, featuring a convolutional layer followed by a max pooling layer. The activity classifier and domain classifier are comprised of a fully connected layer. The non-causal feature extractor and domain classifier are utilized only during training; they would be removed during inference. The primary objective is to evaluate the performance of the proposed domain generalization (DG) algorithm in sensor-based cross-domain activity recognition scenarios, where different subjects utilize the same sensing devices. For all cross-domain tasks, only one target domain is selected for testing in each task. The remaining source domains are used for training, with 20\% of the data left out as a validation set for hyperparameter tuning. The initial learning rates for all algorithms are set at either $10^{-2}$ or $10^{-3}$, and an Adam optimizer is employed for training over 150 epochs with a batch size of 32.

\subsection{The Comparing Baselines}
The proposed causality-inspired learning method is compared with four state-of-the-art HAR approaches: GILE \cite{qian2021latent}, AdaRNN \cite{du2021adarnn}, FIXED \cite{lu2024fixed}, and DIVERSIFY \cite{lu2024diversify}. In addition to the standard ERM \cite{vapnik1991principles} baseline, comparisons are also made with six commonly used domain generalization methods: DANN \cite{ganin2016domain}, CORAL \cite{sun2016deep}, Mixup \cite{zhang2018mixup}, GroupDRO \cite{sagawadistributionally}, RSC \cite{huang2020self}, and ANDMask \cite{parascandololearning}. To ensure fairness, the publicly released code for all comparative baselines was utilized to reproduce results with the same backbone architecture; all methods, except for GILE \footnote{https://github.com/Hangwei12358/cross-person-HAR} and AdaRNN \footnote{https://github.com/jindongwang/transferlearning/tree/master/code/deep/adarnn}, employed identical network architectures. All implementations were conducted in PyTorch and trained on a deep learning server equipped with a RTX 3090 GPU. Experiments were repeated three times using different random seeds, and the average results with 95\% confidence intervals are reported.

\begin{table*}[!t]
	\centering
	\setlength{\tabcolsep}{1.2mm} 
	\caption{Accuracy and Macro F1 scores on cross-person generalization. “Target” 0-3 denotes the unseen test set. For every target domain, the result in the top row is Accuracy, and the result in the bottom row is the Macro F1 score. \textbf{Bold} means the best while \underline{underline} means the second-best.}
	\resizebox{1.0\textwidth}{!}{
		\begin{tabular}{cc|cccccccccccc}
			\toprule
			& Target & $\text{ERM}$ & $\text{DANN}$ & $\text{CORAL}$ & $\text{Mixup}$ & $\text{GroupDRO}$ & $\text{RSC}$ & $\text{ANDMask}$ & $\text{GILE}$ & $\text{AdaRNN}$ & $\text{FIXED}$ &$\text{DIVERSIFY}$ & $\text{Ours}$ \\
			\cmidrule(lr){1-14}
			\multirow{10}{*}{\rotatebox{90}{DSADS}}
			& \multirow{2}{*}{0} & 83.1$\pm$2.7 & 89.1$\pm$2.5 & 91.0$\pm$0.7 & 89.6$\pm$2.5 & \underline{91.7$\pm$0.5} & 84.9$\pm$2.4 & 85.0$\pm$2.3& 81.0$\pm$2.8& 80.9$\pm$2.3 &87.9$\pm$2.2& 90.4$\pm$0.5& \textbf{97.2$\pm$0.2} \\
			&  & 81.9$\pm$2.6 & 87.9$\pm$2.1 & 90.1$\pm$0.8 & 89.1$\pm$2.3 & \underline{90.7$\pm$0.6} & 83.5$\pm$2.1 & 84.3$\pm$2.2& 80.6$\pm$2.6& 80.5$\pm$2.4 &86.8$\pm$2.1& 89.8$\pm$1.2& \textbf{96.5$\pm$0.3} \\ \cmidrule(lr){2-14}
			
			& \multirow{2}{*}{1} & 79.3$\pm$2.5& 84.2$\pm$1.3 & 85.8$\pm$1.5  & 82.2$\pm$2.6  & 85.9$\pm$2.4  & 82.3$\pm$2.1  & 75.8$\pm$2.1  & 75.0$\pm$1.9 & 75.5$\pm$2.3 &79.5$\pm$2.1& \underline{86.5$\pm$2.2} & \textbf{91.3$\pm$0.6} \\
			&  & 78.1$\pm$2.8& 83.8$\pm$1.5 & 85.0$\pm$2.2  & 81.5$\pm$2.7  & 85.5$\pm$2.2  & 82.3$\pm$2.1  & 75.3$\pm$2.2  & 74.6$\pm$2.5 & 75.7$\pm$1.8 &79.1$\pm$2.1& \underline{85.8$\pm$1.7} & \textbf{90.8$\pm$0.7} \\ \cmidrule(lr){2-14}
			
			& \multirow{2}{*}{2} & 87.8$\pm$1.8 & 85.9$\pm$1.6 & 86.6$\pm$2.2  & 89.2$\pm$1.7  & 87.6$\pm$1.8  & 86.7$\pm$1.6  & 87.0$\pm$1.1  & 77.0$\pm$2.1 & \underline{90.2$\pm$1.2} &88.5$\pm$1.8& 90.0$\pm$0.9 & \textbf{93.4$\pm$0.6} \\
			&  & 87.3$\pm$1.8 & 85.6$\pm$2.1 & 86.0$\pm$2.3  & 89.1$\pm$1.6  & 87.1$\pm$1.7  & 86.7$\pm$1.3  & 86.5$\pm$1.2  & 75.8$\pm$1.7 & \underline{89.6$\pm$1.1} &87.6$\pm$1.7& 89.2$\pm$1.3 & \textbf{92.3$\pm$0.5} \\\cmidrule(lr){2-14}
			
			& \multirow{2}{*}{3} & 71.0$\pm$1.3 & 83.4$\pm$1.2 & 78.2$\pm$1.3 & \underline{86.9$\pm$0.8}  & 78.3$\pm$1.4  & 77.7$\pm$1.2  & 77.6$\pm$1.4  & 66.0$\pm$1.9 & 75.5$\pm$1.1 &83.2$\pm$0.9& 86.1$\pm$1.2 & \textbf{93.8$\pm$0.2} \\
			&  & 70.2$\pm$1.6 & 83.5$\pm$1.1 & 77.3$\pm$1.7 & \underline{87.1$\pm$1.5}  & 78.7$\pm$1.2  & 77.9$\pm$1.8  & 77.4$\pm$1.3  & 66.2$\pm$2.1 & 75.0$\pm$1.2 &82.1$\pm$1.1& 85.6$\pm$1.2 & \textbf{93.2$\pm$0.4} \\\cmidrule(lr){2-14}
			
			& \multirow{2}{*}{AVG} & 80.3$\pm$2.2 & 85.6$\pm$1.6 & 85.4$\pm$1.4  & 87.0$\pm$1.9 & 85.9$\pm$1.6 & 82.9$\pm$1.8  & 81.4$\pm$1.6  & 74.7$\pm$2.3 & 80.5$\pm$1.5 &84.8$\pm$1.7& \underline{88.2$\pm$1.3} & \textbf{93.9$\pm$0.4} \\
			&  & 79.4$\pm$2.2 & 85.2$\pm$1.3 & 84.6$\pm$2.0  & 86.7$\pm$2.0 & 85.5$\pm$1.6 & 82.6$\pm$1.7  & 80.1$\pm$1.5  & 74.3$\pm$2.2 & 80.2$\pm$1.3 &83.9$\pm$1.7& \underline{87.6$\pm$1.4} & \textbf{93.2$\pm$0.5} \\
			\midrule
			
			\multirow{10}{*}{\rotatebox{90}{USC-HAD}} 
			& \multirow{2}{*}{0} & 81.0$\pm$1.6 & 81.2$\pm$1.9 & 78.8$\pm$1.7& 80.0$\pm$2.8 & 80.1$\pm$1.3 & 81.9$\pm$1.7 & 79.9$\pm$2.5 & 78.0$\pm$2.8 & 78.6$\pm$2.5 &80.2$\pm$1.1& \underline{82.6$\pm$1.7} & \textbf{84.8$\pm$0.8} \\
			&  & 79.6$\pm$1.8 & 80.3$\pm$1.6 & 78.2$\pm$2.0& 79.3$\pm$2.9 & 80.5$\pm$1.6 & 81.5$\pm$2.0 & 78.5$\pm$2.8 & 77.2$\pm$2.6 & 76.8$\pm$1.8 &79.5$\pm$1.1& \underline{81.6$\pm$1.3} & \textbf{83.5$\pm$0.6} \\\cmidrule(lr){2-14}
			
			& \multirow{2}{*}{1} & 57.7$\pm$2.7 & 57.9$\pm$1.2 & 58.9$\pm$2.1 & \underline{64.1$\pm$2.9} & 55.5$\pm$2.1 & 57.9$\pm$1.1 & 55.3$\pm$2.5 & 62.0$\pm$2.4& 55.3$\pm$2.7 &63.0$\pm$1.3& 63.5$\pm$1.2 & \textbf{66.4$\pm$1.2}  \\
			&  & 56.9$\pm$2.3 & 56.6$\pm$1.6 & 58.3$\pm$1.9 & \underline{63.3$\pm$2.3} & 54.8$\pm$2.1 & 56.6$\pm$0.9 & 53.9$\pm$2.1 & 61.3$\pm$1.9& 54.7$\pm$2.7 &61.8$\pm$1.6& 62.6$\pm$2.1 & \textbf{65.8$\pm$1.1}  \\\cmidrule(lr){2-14}
			
			& \multirow{2}{*}{2} & 74.0$\pm$1.4 & 76.7$\pm$0.9 & 75.0$\pm$1.1 & 74.3$\pm$0.8 & 74.7$\pm$1.1 & 73.4$\pm$2.7 & 74.5$\pm$2.1 & 77.0$\pm$1.3 & 66.9$\pm$1.1 &72.1$\pm$0.8& \textbf{78.7$\pm$1.2} & \underline{78.5$\pm$1.1} \\
			&  & 72.8$\pm$1.7 & 75.8$\pm$1.1 & 74.2$\pm$1.2 & 73.8$\pm$1.0 & 73.9$\pm$0.9 & 72.8$\pm$2.9 & 73.8$\pm$2.1 & 76.2$\pm$1.2 & 65.3$\pm$1.7 &71.2$\pm$1.1& \textbf{80.0$\pm$0.9} & \underline{77.8$\pm$1.3} \\\cmidrule(lr){2-14}
			
			& \multirow{2}{*}{3}  & 65.9$\pm$2.3 & 70.7$\pm$2.1 & 53.7$\pm$1.4 & 61.3$\pm$2.3 & 60.0$\pm$2.5 & 65.1$\pm$2.2 & 65.0$\pm$1.3 & 63.0$\pm$2.8 & \underline{73.7$\pm$1.3} &63.0$\pm$2.8& 71.3$\pm$1.8 & \textbf{76.7$\pm$1.3} \\
			&   & 64.6$\pm$2.6 & 70.1$\pm$2.1 & 52.5$\pm$1.7 & 60.8$\pm$2.2& 59.6$\pm$2.3 & 64.3$\pm$1.9 & 63.8$\pm$1.5 & 61.8$\pm$2.8 & \underline{72.9$\pm$0.9} &62.3$\pm$2.1& 70.6$\pm$2.2 & \textbf{76.1$\pm$1.5} \\\cmidrule(lr){2-14}
			
			& \multirow{2}{*}{AVG}  & 69.7$\pm$2.0 & 71.6$\pm$1.3 & 66.6$\pm$1.8 & 69.9$\pm$2.6 & 67.6$\pm$1.7 & 69.6$\pm$2.0 & 68.7$\pm$2.2 & 70.0$\pm$2.5 & 68.6$\pm$2.1 &69.6$\pm$1.5& \underline{74.0$\pm$1.6} & \textbf{76.6$\pm$1.0} \\
			&   & 68.5$\pm$1.9 & 70.7$\pm$1.5 & 65.8$\pm$1.6 & 69.3$\pm$2.2 & 67.2$\pm$1.7 & 68.8$\pm$2.1 & 67.5$\pm$2.2 & 69.1$\pm$2.3 & 67.4$\pm$1.8 &68.7$\pm$1.3& \underline{73.2$\pm$1.5} & \textbf{75.8$\pm$1.2} \\
			\midrule
			
			\multirow{10}{*}{\rotatebox{90}{PAMAP2}} 
			& \multirow{2}{*}{0} & 90.0$\pm$0.7 & 82.2$\pm$1.2 & 86.2$\pm$1.0 & 89.4$\pm$0.9 & 85.2$\pm$1.8 & 87.1$\pm$1.6 & 86.7$\pm$0.8 & 83.0$\pm$1.6 & 81.6$\pm$1.2 &90.7$\pm$0.8& \underline{91.0$\pm$0.6} & \textbf{93.2$\pm$0.4}\\
			&  & 89.2$\pm$1.3 & 81.3$\pm$1.0 & 84.8$\pm$1.3 & 88.2$\pm$1.1 & 83.9$\pm$1.7 & 85.7$\pm$1.8 & 85.9$\pm$1.1 & 82.3$\pm$2.0 & 80.4$\pm$1.0 &90.1$\pm$0.6& \underline{90.2$\pm$0.9} & \textbf{92.8$\pm$0.5}\\\cmidrule(lr){2-14}
			
			& \multirow{2}{*}{1} & 78.1$\pm$2.6 & 78.1$\pm$2.8 & 77.8$\pm$1.1 & 80.3$\pm$2.7 & 77.7$\pm$3.1 & 76.9$\pm$3.1 & 76.4$\pm$2.3 & 68.0$\pm$3.2 & 71.8$\pm$2.5 &81.2$\pm$2.5& \underline{84.3$\pm$2.1}& \textbf{88.6$\pm$1.5}  \\
			&  & 77.4$\pm$2.3 & 77.3$\pm$2.4 & 77.1$\pm$1.5 & 79.6$\pm$2.5 & 76.4$\pm$2.7 & 75.3$\pm$2.8 & 75.1$\pm$2.5 & 66.8$\pm$2.9 & 70.4$\pm$3.2 &80.3$\pm$2.2& \underline{83.1$\pm$1.7}& \textbf{87.8$\pm$1.2}  \\ \cmidrule(lr){2-14}
			
			& \multirow{2}{*}{2} & 55.8$\pm$2.7 & 55.8$\pm$2.6 & 49.0$\pm$2.3 & 58.4$\pm$2.6 & 56.2$\pm$2.6 & 60.3$\pm$2.4& 43.6$\pm$2.3 & 42.0$\pm$1.3 & 45.4$\pm$2.5 &54.2$\pm$2.7& \underline{60.5$\pm$1.1} & \textbf{62.8$\pm$1.2} \\
			&  & 54.2$\pm$2.2 & 54.6$\pm$2.8 & 48.2$\pm$2.7 & 57.8$\pm$2.8 & 55.4$\pm$2.5 & 59.1$\pm$2.6& 43.2$\pm$2.1 & 41.4$\pm$1.7 & 44.2$\pm$2.7 &53.4$\pm$2.6& \underline{59.6$\pm$1.6} & \textbf{62.2$\pm$1.3} \\ \cmidrule(lr){2-14}
			
			& \multirow{2}{*}{3}  & 84.4$\pm$1.4 & 87.3$\pm$1.1 & 87.8$\pm$1.6 & 87.7$\pm$2.8 & 85.0$\pm$2.1 & 87.8$\pm$1.3 & 85.6$\pm$1.4 & 76.0$\pm$1.2 & 82.7$\pm$1.7&\underline{90.7$\pm$0.7}& 87.7$\pm$2.0 & \textbf{91.8$\pm$0.5} \\
			&   & 83.2$\pm$1.7 & 86.5$\pm$1.4 & 86.3$\pm$2.0 & 87.1$\pm$2.3 & 83.8$\pm$1.9 & 86.7$\pm$1.8 & 84.2$\pm$1.9 & 74.8$\pm$1.8 & 81.3$\pm$2.0&\underline{89.8$\pm$0.9}& 86.4$\pm$1.8 & \textbf{91.2$\pm$0.7} \\ \cmidrule(lr){2-14}
			
			& \multirow{2}{*}{AVG}  & 77.1$\pm$1.7 & 75.7 $\pm$2.0& 75.2$\pm$1.2 & 79.0$\pm$2.5 & 76.0$\pm$2.2 & 78.0$\pm$2.0& 73.1$\pm$1.7 & 67.5$\pm$2.2 & 70.4$\pm$2.0 &79.2$\pm$1.8& \underline{80.8$\pm$1.3} & \textbf{84.1$\pm$0.9} \\
			&  & 76.0$\pm$1.9 & 74.9 $\pm$1.7& 74.1$\pm$1.6 & 78.2$\pm$2.2 & 74.9$\pm$2.1 & 76.7$\pm$2.2& 72.1$\pm$2.0 & 66.3$\pm$1.9 & 69.0$\pm$2.3 &78.4$\pm$2.0& \underline{79.8$\pm$1.4} & \textbf{83.5$\pm$1.1} \\ \cmidrule(lr){1-14}
			
			& AVG all & 75.7 & 77.7 & 75.7 & 78.6 & 76.5 & 76.9 & 74.4 & 70.7 & 73.2 &77.9& \underline{81.0} & \textbf{84.9} \\
			\bottomrule
		\end{tabular}
		\label{tab_crpe}
	}
\end{table*}

\begin{table*}[t]
	\small
	\centering
	\setlength{\tabcolsep}{1.2mm}
	\caption{Accuracy and Macro F1 scores on cross-position generalization. “Target” 0-4 denotes the unseen test set. For every target domain, the result in the top row is Accuracy, and the result in the bottom row is the Macro F1 score. \textbf{Bold} means the best while \underline{underline} means the second-best.}
	\resizebox{1.0\textwidth}{!}{
		\begin{tabular}{cc|cccccccccc}
				\toprule
				&Target & $\text{ERM}$ & $\text{DANN}$ & $\text{CORAL}$ & $\text{Mixup}$ & $\text{GroupDRO}$ & $\text{RSC}$ & $\text{ANDMask}$ & $\text{FIXED}$ & $\text{DIVERSIFY}$ & $\text{Ours}$ \\
				\cmidrule(lr){1-12}
				\multirow{12}{*}{\rotatebox{90}{DSADS}}
					&\multirow{2}{*}{0} & 41.5$\pm$3.2 & 45.4$\pm$2.2 & 33.2$\pm$2.8   & \underline{48.8$\pm$2.8}  & 27.1$\pm$2.3  & 46.6$\pm$2.5  & 47.5$\pm$3.7 & 46.9$\pm$2.8 & 47.7$\pm$2.2 & \textbf{50.2$\pm$1.2} \\
					& & 36.8$\pm$2.7 & 38.5$\pm$2.5 & 27.6$\pm$3.0   & \underline{42.4$\pm$2.5}  & 22.3$\pm$2.5  & 40.6$\pm$3.2  & 41.5$\pm$3.2 & 40.2$\pm$3.2 & 42.2$\pm$2.5 & \textbf{46.2$\pm$1.2} \\ \cmidrule(lr){2-12}
					
					&\multirow{2}{*}{1} & 26.7$\pm$4.5 & 25.3$\pm$3.5 & 25.2$\pm$3.1  & \underline{34.2$\pm$3.2}  & 26.7$\pm$1.8  & 27.4$\pm$4.1  & 31.1$\pm$4.8   &31.6$\pm$3.6& 32.9$\pm$2.9 & \textbf{36.5$\pm$2.3} \\
					& & 24.2$\pm$3.5 & 21.3$\pm$3.2 & 20.8$\pm$3.2  & \underline{31.2$\pm$3.6}  & 22.9$\pm$2.2  & 28.1$\pm$3.6  & 24.1$\pm$4.3   &28.3$\pm$3.8& 27.1$\pm$4.1 & \textbf{31.8$\pm$2.5} \\ \cmidrule(lr){2-12}
					
					&\multirow{2}{*}{2} & 35.8$\pm$3.6 & 38.1$\pm$2.4 & 25.8$\pm$3.8  & 37.5$\pm$3.4  & 24.3$\pm$2.4  & 35.9$\pm$3.2  & 39.2$\pm$2.3   &41.5$\pm$1.5& \textbf{44.5$\pm$1.6} & \underline{43.8$\pm$1.5} \\
					& & 31.2$\pm$4.6 & 33.5$\pm$2.3 & 20.2$\pm$3.7  & 35.1$\pm$3.2  & 21.2$\pm$2.7  & 30.5$\pm$2.9  & 33.8$\pm$2.6   &36.2$\pm$1.7& \textbf{39.8$\pm$1.9} & \underline{40.2$\pm$1.6} \\ \cmidrule(lr){2-12}
					
					&\multirow{2}{*}{3} & 21.4$\pm$4.2 & 28.9$\pm$3.9 & 22.3$\pm$2.4  & 29.5$\pm$4.5  & 18.4$\pm$4.8  & 27.0$\pm$2.5  & 30.2$\pm$3.4 &28.6$\pm$2.7& \underline{31.6$\pm$3.2} & \textbf{31.8$\pm$1.7} \\
					& & 14.8$\pm$4.5 & 22.6$\pm$4.2 & 17.3$\pm$3.5  & 24.8$\pm$3.2  & 14.2$\pm$4.5  & 21.2$\pm$2.7  & 24.8$\pm$3.2 &20.8$\pm$3.1& \underline{27.8$\pm$2.8} & \textbf{28.3$\pm$1.8} \\ \cmidrule(lr){2-12}
					
					&\multirow{2}{*}{4} & 27.3$\pm$2.7 & 25.1$\pm$2.3 & 20.6$\pm$4.0  & 29.9$\pm$5.2  & 24.8$\pm$3.5  & 29.8$\pm$2.2  & 29.9$\pm$1.8   &29.6$\pm$1.8& \underline{30.4$\pm$2.8} & \textbf{33.2$\pm$3.2} \\
					& & 24.6$\pm$2.3 & 21.2$\pm$2.1 & 16.2$\pm$3.8  & 27.7$\pm$4.5  & 18.6$\pm$3.7  & 21.4$\pm$2.5  & 24.3$\pm$2.4   &25.8$\pm$2.2& \underline{26.8$\pm$3.2} & \textbf{28.6$\pm$2.8} \\ \cmidrule(lr){2-12}
					
					&\multirow{2}{*}{AVG} & 30.6$\pm$3.8 & 32.6$\pm$3.0 & 25.4$\pm$3.4  & 36.0$\pm$4.2 & 24.3$\pm$3.3 & 33.3$\pm$2.7  & 35.6$\pm$3.5   & 35.6$\pm$2.3 & \underline{37.4$\pm$2.6} & \textbf{39.1$\pm$1.8} \\
					& & 26.3$\pm$3.6 & 27.4$\pm$2.8 & 20.4$\pm$3.3  & 32.2$\pm$3.4 & 19.8$\pm$2.9 & 28.4$\pm$3.0  & 29.7$\pm$3.1   & 30.3$\pm$2.6 & \underline{32.7$\pm$2.7} & \textbf{35.0$\pm$2.0} \\ 
					\bottomrule
			\end{tabular}
		\label{tab_crpo}
	    }
\end{table*}

\begin{table*}[!t]
	\centering
	\setlength{\tabcolsep}{1.2mm}
	\caption{Accuracy and Macro F1 scores on cross-dataset generalization. “Target” 0-3 denotes the unseen test set. For every target domain, the result in the top row is Accuracy, and the result in the bottom row is the Macro F1 score. \textbf{Bold} means the best while \underline{underline} means the second-best.}
	\resizebox{1.0\textwidth}{!}{
		\begin{tabular}{c|cccccccccc}
			\toprule
			Target & $\text{ERM}$ & $\text{DANN}$ & $\text{CORAL}$ & $\text{Mixup}$ & $\text{GroupDRO}$ & $\text{RSC}$ & $\text{ANDMask}$ & $\text{FIXED}$ & $\text{DIVERSIFY}$ & $\text{Ours}$ \\
			\cmidrule(lr){1-11}
			\multirow{2}{*}{0} & 26.4$\pm$2.3 & 29.7$\pm$3.4 & 39.5$\pm$2.5  & 37.3$\pm$2.2& \underline{51.4$\pm$1.6}  & 33.1$\pm$2.8  & 41.7$\pm$2.3 &49.6$\pm$1.6& 48.7$\pm$2.3 & \textbf{52.6$\pm$1.7}  \\
			& 20.6$\pm$4.2 & 18.2$\pm$2.5 & 26.2$\pm$2.2  & 23.1$\pm$2.9& 42.2$\pm$2.2  & 22.7$\pm$4.2  & 40.5$\pm$2.1 &\underline{43.2$\pm$2.1}& 41.6$\pm$2.6 & \textbf{43.2$\pm$1.8}  \\\cmidrule(lr){1-11}
			
			\multirow{2}{*}{1} & 29.6$\pm$3.4 & 45.3$\pm$2.5 & 41.8$\pm$3.1 & \underline{47.4$\pm$2.4}  & 36.7$\pm$2.4  & 39.7$\pm$3.1  & 33.8$\pm$2.5  & 46.8$\pm$2.1& 46.9$\pm$3.1 & \textbf{49.2$\pm$2.5} \\
			& 23.1$\pm$3.5 & 39.8$\pm$3.5 & 27.8$\pm$2.4 & \underline{42.6$\pm$2.1}  & 23.4$\pm$3.4  & 33.6$\pm$3.7  & 22.6$\pm$3.6  & 40.2$\pm$2.6& 42.3$\pm$3.1 & \textbf{45.1$\pm$2.1} \\\cmidrule(lr){1-11}
			
			\multirow{2}{*}{2} & 44.4$\pm$2.4 & 46.1$\pm$3.2 & 39.1$\pm$2.4  & 40.2$\pm$3.1  & 33.2$\pm$3.2  & 45.3$\pm$2.4  & 43.2$\pm$3.1   &47.2$\pm$2.4& \underline{49.0$\pm$2.3} & \textbf{49.8$\pm$1.2} \\
			& 38.1$\pm$2.2 & 41.4$\pm$2.8 & 32.5$\pm$3.1  & 32.8$\pm$2.5  & 25.6$\pm$3.5  & 36.9$\pm$2.4  & 33.5$\pm$2.8   &\underline{41.8$\pm$1.8}& 39.0$\pm$1.6 & \textbf{42.5$\pm$1.5} \\\cmidrule(lr){1-11}
			
			\multirow{2}{*}{3} & 32.9$\pm$3.2 & 43.8$\pm$2.5 & 36.6$\pm$3.6  & 23.1$\pm$4.2  & 33.8$\pm$3.5  & 45.9$\pm$2.4  & 40.2$\pm$2.8   &52.4$\pm$1.5& \underline{59.9$\pm$1.7} & \textbf{60.8$\pm$0.8}  \\
			& 28.6$\pm$2.9 & 39.8$\pm$2.2 & 29.6$\pm$2.9  & 20.7$\pm$3.2  & 26.4$\pm$2.6  & 39.6$\pm$2.9  & 36.5$\pm$3.2   &42.7$\pm$2.5& \underline{48.6$\pm$1.8} & \textbf{51.6$\pm$1.4}  \\\cmidrule(lr){1-11}
			
			\multirow{2}{*}{AVG} & 33.3$\pm$2.5 & 41.2$\pm$2.7 & 39.2$\pm$2.8  & 37.0$\pm$3.2 & 38.8$\pm$2.7 & 41.0$\pm$2.7  & 39.7$\pm$2.5   &49.0$\pm$1.8& \underline{51.1$\pm$2.5} & \textbf{53.1$\pm$1.5}  \\
			& 27.6$\pm$3.2 & 34.8$\pm$2.7 & 29.0$\pm$2.5  & 29.8$\pm$2.7 & 29.4$\pm$3.0 & 33.2$\pm$3.2  & 33.3$\pm$3.1   &42.0$\pm$2.2& \underline{42.9$\pm$2.4} & \textbf{45.6$\pm$1.7}  \\
			\bottomrule
		\end{tabular}
		\label{tab_crda}
	}
\end{table*}

\begin{table*}[!t]
	\centering
	\setlength{\tabcolsep}{1.2mm}
	\small
	\caption{Accuracy and Macro F1 scores on one-person-to-another generalization. For every target domain, the result in the top row is Accuracy, and the result in the bottom row is the Macro F1 score. \textbf{Bold} means the best while \underline{underline} means the second-best.}
	\begin{tabular}{l|cccccccc}
			\toprule
			Target & $\text{ERM}$ & $\text{Mixup}$ & $\text{GroupDRO}$ & $\text{RSC}$ & $\text{ANDMask}$ & $\text{FIXED}$ & $\text{DIVERSIFY}$ & $\text{Ours}$ \\
			\cmidrule(lr){1-9}
			\multirow{2}{*}{DSADS} & 51.3$\pm$2.3 & 62.7$\pm$1.7 & 51.3$\pm$2.1  & 59.1$\pm$1.6  & 57.2$\pm$1.6 &60.8$\pm$1.3& \underline{67.6$\pm$1.1} & \textbf{71.5$\pm$0.9}  \\
			& 47.1$\pm$2.8 & 57.2$\pm$1.4 & 47.5$\pm$2.5  & 55.5$\pm$2.1  & 53.5$\pm$1.8 &56.4$\pm$1.1& \underline{63.3$\pm$1.3} & \textbf{67.5$\pm$0.6}  \\\cmidrule(lr){1-9}
			
			\multirow{2}{*}{USC-HAD} & 46.2$\pm$2.5  & 46.3$\pm$2.8  & 48.0$\pm$2.3  & 49.0$\pm$1.9  & 45.9$\pm$2.1 &53.1$\pm$1.9& \underline{55.0$\pm$1.5} & \textbf{58.2$\pm$1.3} \\
			& 42.5$\pm$1.8  & 41.2$\pm$2.2  & 44.5$\pm$2.6  & 45.8$\pm$2.4  & 42.1$\pm$2.1 &47.9$\pm$2.6& \underline{50.7$\pm$1.8} & \textbf{53.6$\pm$1.5} \\\cmidrule(lr){1-9}
			
			\multirow{2}{*}{PAMAP2} & 53.1$\pm$1.6 & 58.6$\pm$1.5  & 53.1$\pm$1.7  & 59.7$\pm$1.3  & 54.3$\pm$1.5   &59.8$\pm$1.4& \underline{62.5$\pm$1.2} & \textbf{63.2$\pm$0.8} \\
			& 48.9$\pm$2.1 & 53.5$\pm$2.3  & 49.6$\pm$2.3  & 54.5$\pm$2.1  & 51.2$\pm$1.8   &55.3$\pm$1.6& \underline{58.8$\pm$1.4} & \textbf{59.6$\pm$1.2} \\\cmidrule(lr){1-9}
			
			\multirow{2}{*}{AVG} & 50.2$\pm$2.2   & 55.8$\pm$2.0 & 50.8$\pm$1.9 & 55.9$\pm$1.5  & 52.5$\pm$1.8   &57.7$\pm$1.6& \underline{61.7$\pm$1.3} & \textbf{64.3$\pm$1.0}  \\
			& 46.2$\pm$2.0   & 50.6$\pm$1.8 & 47.2$\pm$2.4 & 51.9$\pm$2.2  & 48.9$\pm$1.9   &53.2$\pm$1.8& \underline{57.6$\pm$1.5} & \textbf{60.2$\pm$1.1}  \\
			\bottomrule
		\end{tabular}
	\label{tab_one}
\end{table*}

\subsection{Main Results}
We report our main results according to classification accuracy and macro F1 score. \tablename~\ref{tab_crpe} to \ref{tab_one} summarize our key findings, from which four conclusions can be drawn: (1) \tablename~\ref{tab_crpe} presents results from cross-subject generalization settings. In this context, the ERM baseline method performs comparably to some current approaches and even outperforms others on specific tasks, such as DANN, CORAL, GroupDRO, and AdaRNN on the PAMAP2 dataset. This observation aligns closely with previous findings \cite{gulrajani2021search}, likely due to these methods' inability to effectively reduce the distributional differences in time-series sensor data. Consequently, it is crucial to explore causal features that can effectively mitigate these distributional discrepancies in human activity recognition (HAR) sensor data. (2) Overall, our proposed causal-inspired method consistently demonstrates superior performance compared to other state-of-the-art baselines in cross-subject settings, achieving optimal results across three benchmark datasets. For instance, in terms of average accuracy, our method exceeds the current SOTA method, DIVERSIFY, by 5.7\%, 2.6\%, and 3.3\% on the DSADS, USC-HAD, and PAMAP datasets, respectively. (3) As noted, other methods, such as Mixup, achieve suboptimal accuracies on DSADS, but only marginally outperform ERM by 0.2\% on USC-HAD. Methods like CORAL, GroupDRO, and ANDMask even perform below the ERM baseline on USC-HAD. This phenomenon may stem from their disregard for intrinsic causal mechanism, potentially resulting in entangled activity semantic and domain-specific features across different distributions. (4) As illustrated in \tablename~\ref{tab_crpo} to \ref{tab_one}, our method consistently achieves comparable or superior performance under cross-position, cross-dataset, and one-person-to-another settings. For example, cross-position scenarios are known to be more challenging than the other two cases. In this context, our causality-inspired method exceeds the best baseline by 1.7\% in terms of average accuracy, while also achieving approximately an 8.5\% accuracy improvement compared to ERM in cross-position settings. These results indicate that our method possesses strong generalization capabilities for time-series classification across various domain generalization evaluation settings.

\begin{figure}[!t]
	\centering
	\subfigure[DSADS]{
		\includegraphics[width=0.42\linewidth]{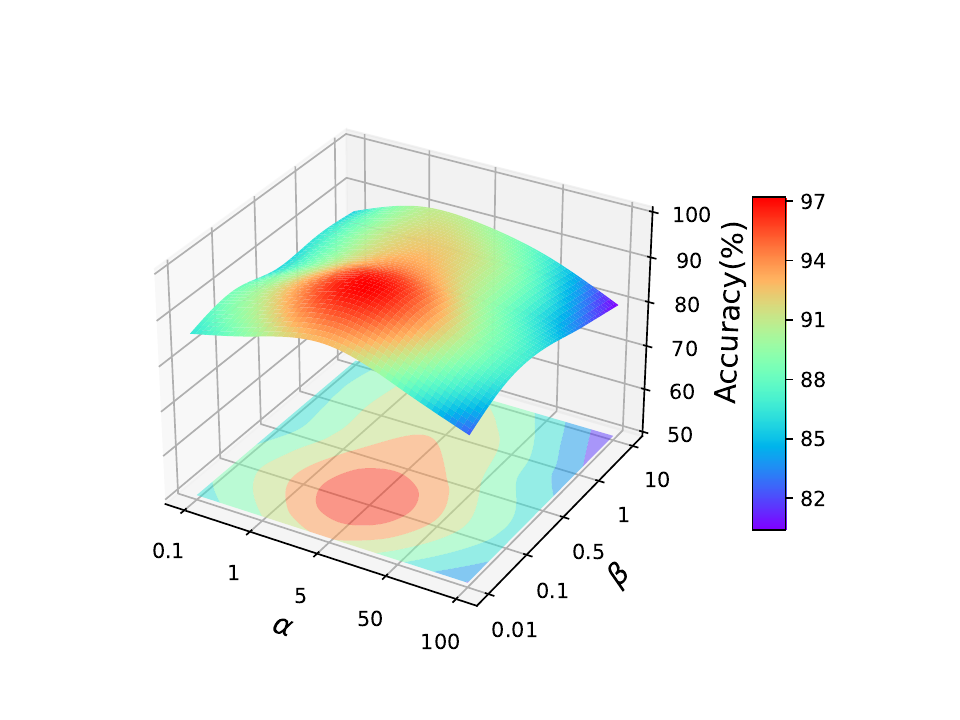}
		\label{fig:subfig_ps_1}
	}
	\subfigure[USC-HAD]{
		\includegraphics[width=0.42\linewidth]{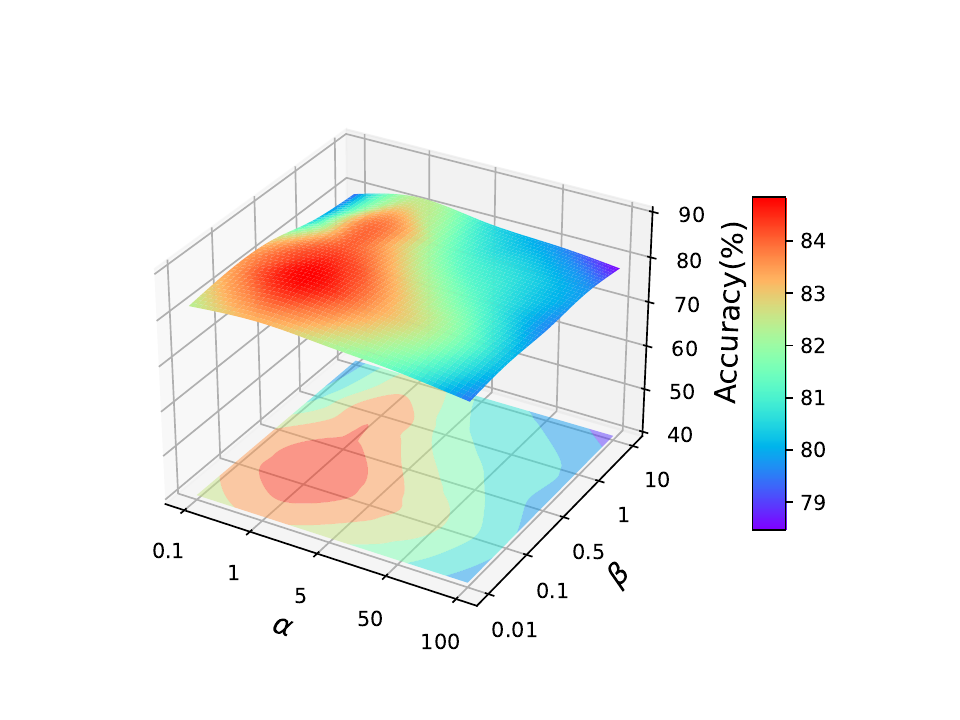}
		\label{fig:subfig_ps_2}
	}
	\vspace{-.2in}
	\caption{Parameters sensitivity analysis of $\alpha$ and $\beta$. \textbf{(a)} The experiment is conducted on the DSADS dataset, with "0" designated as the target domain. \textbf{(b)} The experiment is conducted on the USC-HAD dataset, also using "0" as the target domain.}
	\label{fig_ps}
\end{figure}

\subsection{Parameter Sensitivity Analysis}
In our causality-inspired learning approach, there are two main hyper-parameters, i.e., $\alpha$ and $\beta$ contained in our total learning objective, which are in charge of controlling the loss intensity of HSIC and IDS, respectively, as previously depicted in Eq. (\ref{eq_9}). To analyze their sensitivity, we also perform an ablation study through evaluating our method on the DSADS and USC-HAD datasets under different hyper-parameter ranges of them. Specifically, as a linear combination of the two loss terms, the value of $\alpha$ changes from $\{0.1, 1, 5, 50,100\}$, while $\beta$ varies from $\{0.01, 0.1, 0.5, 1, 10\}$. The results are shown in \figurename~\ref{fig_ps}. As illustrated in these figures, we can observe dramatic performance drops when either $\alpha$ and $\beta$ is set to be too large, indicating that every loss term plays an essential role in improving domain generalization. While both of them are well balanced for disentangling the causal and non-causal features for cross-domain activity recognition, it can be seen that our method may maintain competitive performance stalely across a broad range of their hyper-parameter values, which further validate the robustness of our method to the variations of two hyper-parameters $\alpha$ and $\beta$.

\subsection{Ablation} \label{sec:ablation}
\begin{table*}[!t]
	\centering
	\caption{The ablations with classification accuracy on the loss term, the HSIC-based independence constraint, the IDS augmentation strategy, CDPL, and the Early-Forking Two-Branch framework assess their indepedent effectiveness using the DSADS dataset under cross-person setting.}
	\begin{tabular}{@{}llcccccc@{}}
		\toprule
		&\multirow{2}{*}{Model} &  \multicolumn{4}{c}{Test Domain} & \multirow{2}{*}{Average} \\\cmidrule(lr){3-6}
		&&  0 & 1 & 2& 3& \\ \midrule
		&Ours  &\textbf{97.2} &\textbf{91.3} &\textbf{93.4} &\textbf{93.8} &\textbf{93.9} \\ \midrule
		\multirow{3}{*}{Loss Terms}&Ours w/o $\mathcal{L}_\text{ind}$ &88.6 &87.2 &86.7 &85.5 &87.0 \\
		&Ours w/o $\mathcal{L}_\text{con}$ &92.6 &\underline{89.6} &90.8 &90.6 &90.9 \\
		&Ours w/o $\mathcal{L}_\text{ind}$ \& $\mathcal{L}_\text{con} $ &86.2 &84.5 &88.3 &79.8 &84.7 \\
		\midrule
		\multirow{2}{*}{Measurements} &Ours w/o $\mathcal{L}_\text{ind}$ \& w/ $\mathcal{L}_\text{orth}$  &94.5 &89.4 &91.2 &90.5 &91.4 \\
		&Ours w/o $\mathcal{L}_\text{ind}$ \& w/ $\mathcal{L}_\text{corr}$  &93.2 &88.7 &90.4 &89.7 &90.5 \\ \midrule
		\multirow{1}{*}{Projection MLP layer}&Ours w/o CDPL &\underline{95.6} &88.3 &\underline{91.6} &91.1 &\underline{91.7}\\ \midrule
		\multirow{3}{*}{Augmentations} 
		&Ours w/o IDS \& w/ MixStyle &92.8 &88.9 &90.2 &89.3 &90.3\\
		&Ours w/o IDS \& w/ DSU &93.4 &88.3 &89.2 &\underline{91.5} &90.6\\
		&Ours w/o IDS &89.1 & 89.2 &88.9 &86.5 &88.4 \\ \midrule
		\multirow{1}{*}{Early-Forking}&Ours w/o $f_c$ / $f_d$&89.1 &89.3 &88.9 &86.7 &88.5 \\
		
		\bottomrule
	\end{tabular}%
	\vspace{-.1in}
	\label{tab_eff}
\end{table*}

\begin{figure}[!t]
	\centering
	\subfigure[Loss Terms]{
		\includegraphics[width=0.23\linewidth]{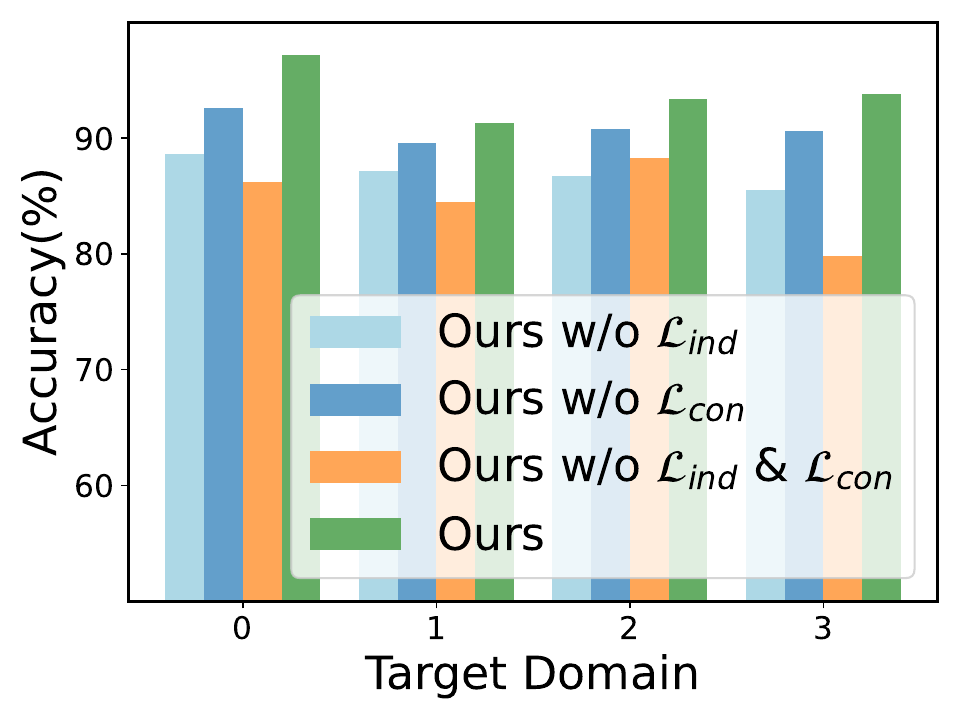}
		\label{fig:ablation_loss_term}
	}
	\subfigure[Measurements]{
		\includegraphics[width=0.23\linewidth]{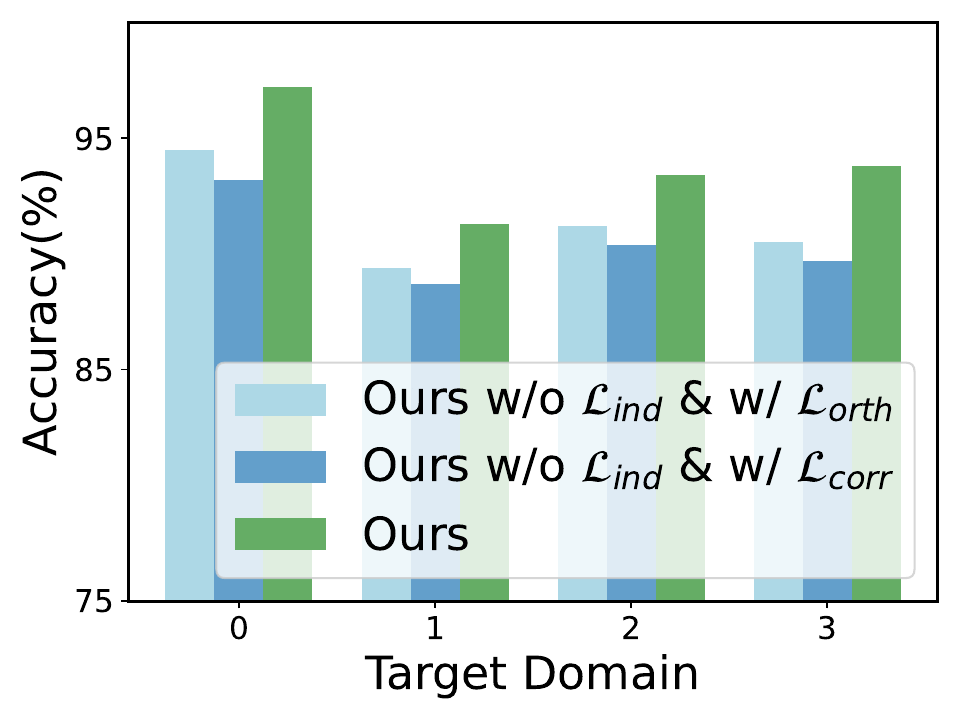}
		\label{fig:ablation_measure}
	}
	\subfigure[CDPL and Early-Forking]{
		\includegraphics[width=0.23\linewidth]{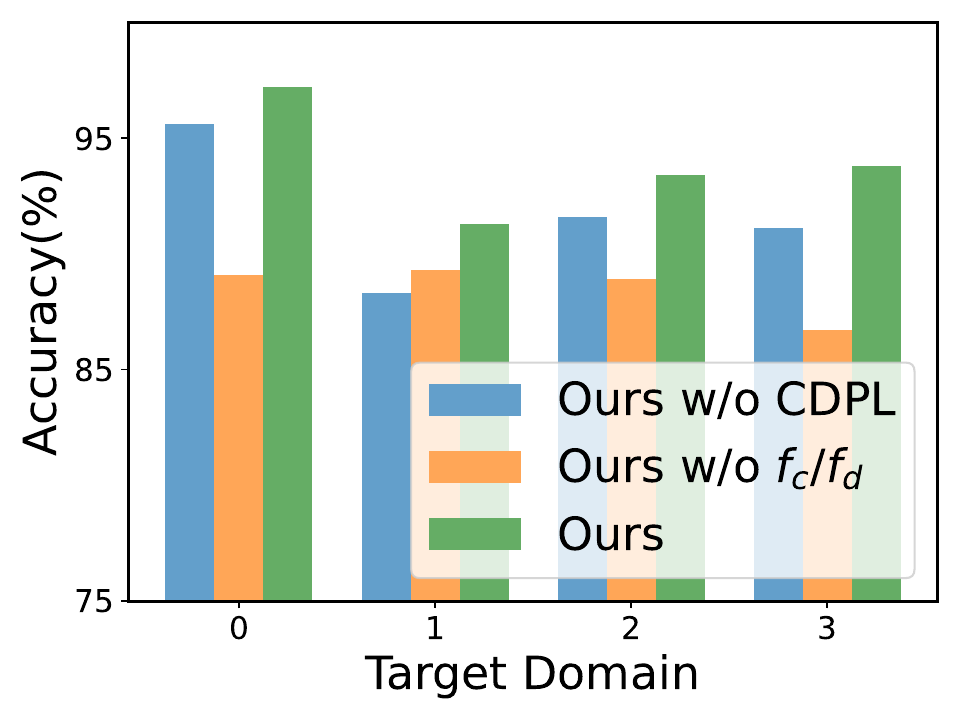}
		\label{fig:ablation_fork_cdpl}
	}
	\subfigure[Augmentation]{
		\includegraphics[width=0.23\linewidth]{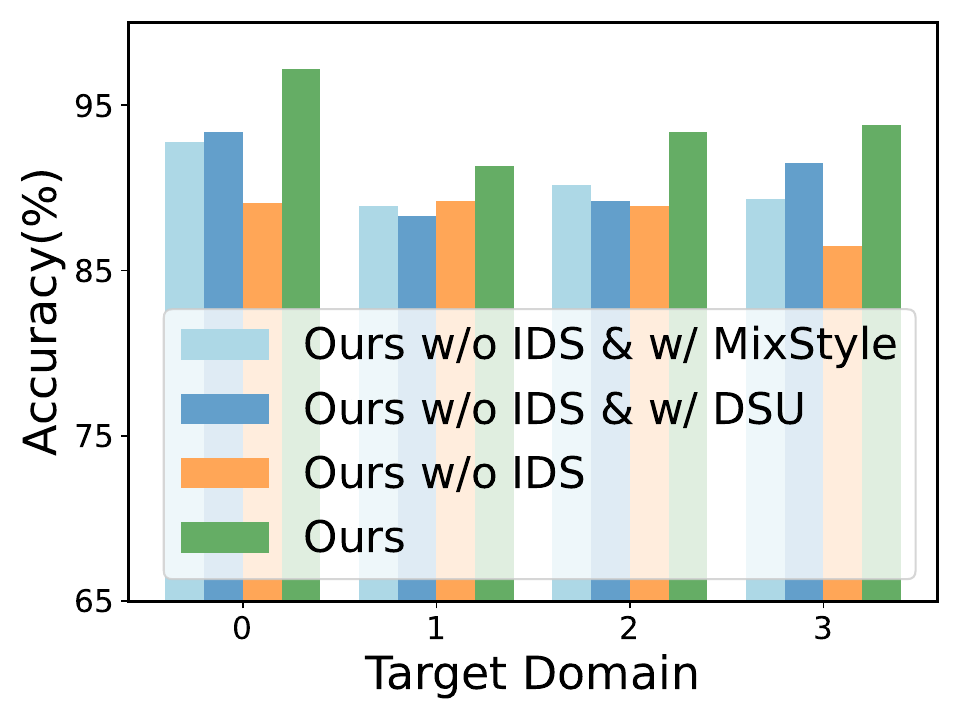}
		\label{fig:ablation_augmentaion}
	}
	\vspace{-.20in}
	\caption{The ablations on the loss term, the HSIC-based independence constraint, the IDS augmentation strategy, CDPL, and the Early-Forking Two-Branch framework assess their indepedent effectiveness using the DSADS dataset under cross-person setting.}
	\label{fig_ablation}
\end{figure}

We explore the independent effect of each module in our causality-motivated representation learning algorithm, as shown in \tablename~\ref{tab_eff} and \figurename~\ref{fig_ablation}. More details are introduced as following.

\textbf{The proposed  constraints in feature consistency and disentanglement are essential for DG-HAR.}
As previously depicted in Eq. (\ref{eq_9}), besides the main cross-entropy losses (i.e., $\mathcal{L}_\text{cls}$), there exist both of additional loss terms, including the independence constraint loss $\mathcal{L}_\text{ind}$ and the conditional consistency loss $\mathcal{L}_\text{con}$. We first ablate their independent effect of two additional loss terms on the DSADS dataset, through separately disabling either or both of them in our complete loss term. From the ablation results summarized in the 2th, 3th and 4th rows of \tablename~\ref{tab_eff}, in comparison with our original loss term as illustrated in Eq. (\ref{eq_9}), it can be seen that disabling either of the two additional loss terms would result in a considerable accuracy decrease (6.9\% and 3.0\%), suggesting that every loss term in our design should have its independent contribution to the final performance improvement. On the other hand, while disabling both of the two loss terms at the same time, we can observe that the resulted model performs the worst, which is inferior to ours as well as its two variants, indicating both of the additional loss terms could be complementary in our design, while further improving its generalization ability.

\textbf{HSIC outperforms other independence metrics.}
Since our main goal is to decease the dependence between the causal (i.e., $F_c(x)$) and non-casual features (i.e., $F_d(x)$) by minimizing the HSIC value between both of them, it is especially crucial to check whether HSIC can serve as a promising solution, while comparing against other two existing popular feature independence metrics, based on orthogonality and correlation, respectively. To clarify this problem, we conduct an ablation study on the DSADS dataset through replacing such HSIC criteria with the other two metrics, where the former (i.e., the correlation constraint denoted as $\mathcal{L}_\text{corr}$) tends to minimize the cross-correlation matrix calculated between $F_c(x)$ and $F_d(x)$, while the latter (i.e., the orthogonal constraint denoted as $\mathcal{L}_\text{orth}$) requires the multiplication of the two features to be zero. We adopt the implementations from \cite{zbontar2021barlow,ranasinghe2021orthogonal} for this ablation study (i.e. Ours w/ $\mathcal{L}_\text{corr}$ or ours w/ $\mathcal{L}_\text{orth}$). As listed in the 5th and 6th rows of \tablename~\ref{tab_eff}, it can be seen that the employed HSIC independence constraint considerably outperforms the other two metrics by 2.5\% and 3.4\% respectively in terms of average accuracy, offering a more effective solution. Therefore, we always apply such HSIC independence criterion to disentangle the casual and non-casual features through the whole paper.

\textbf{The additional CDPL effectively mitigates representation collapse.}
Borrowing a recent idea that attachs an additional projection MLP layer to prevent representation collapse in a two-branch framework \cite{chen2021exploring}, we also perform an ablation study on the DSADS dataset to check whether it does really take into effect in our two-branch design. To achieve this goal, we compare our proposed approach with its variant through removing the attached projection MLP layer in Eq. (\ref{eq_6}). From the comparison results presented in the 7th row of \tablename~\ref{tab_eff}, we can see that the variant, i.e., ‘Ours w/o CDPL’that disables the projection MLP layer is obviously inferior to ours in all test domains, and results in a considerable performance decrease of 2.2\% in terms of average accuracy, thereby verifying its effectiveness in improving such cross-domain activity recognition performance.
\textbf{IDS is significantly superior to other sensor data augmentation in domain style transfer.}
As a data augmentation strategy, our IDS plays a crucial role in altering the domain style of a given sensor sample, while keeping its activity semantic information consistent. To verify its effectiveness, we further compare our method against the baseline without incorporating any other data augmentation scheme. Results are listed in the 10th row of \tablename~\ref{tab_eff}. While allowing data augmentation, it can be seen that our proposed approach can consistently beat the comparing baseline (i.e., Ours w/o augmentation) by a clear margin in all test domains. In addition, we further compare our IDS (i.e., inhomogeneous domain sampling) augmentation strategy with its two variants including MixStyle \footnote{https://github.com/KaiyangZhou/mixstyle-release} \cite{zhoudomain} and DSU \footnote{https://github.com/lixiaotong97/DSU} \cite{li2022uncertainty}, which directly modify $\mu$ and $\sigma^2$ by mixing with other domain styles at instance level and feature level, respectively. As claimed in previous literatures \cite{zhoudomain,li2022uncertainty}, the feature statistics (i.e., mean and standard deviation) usually carry the domain characteristics of training data, which may be properly manipulated to enhance model generalization ability. Both MixStyle and DSU can be viewed as two popular feature/style-based augmentation schemes for domain generalization (DG), which aim to generate new domain styles during training. In the former case (i.e., Ours w/MixStyle \cite{zhoudomain}), MixStyle generates new domain styles by randomly selecting two samples from different domains and performing a probabilistic convex combination between their instance-level feature statistics of bottom CNN layers. In the latter case (i.e., Ours w/DSU \cite{li2022uncertainty}), DSU aims to model the uncertainty of potential domain shifts with synthesized feature statistics. To this end, it hypothesizes that the feature statistic, i.e., mean and standard deviation are not deterministic values, which instead follow a multivariate Gaussian distribution. It may generate new domain styles at feature level by randomly sampling the feature statistics from the estimated Gaussian distribution. Then the synthetized feature statistics variants can be utilized to produce diverse domain shifts, improving the model robustness against potential statistics shifts. From the results listed in the 8th and 9th rows of \tablename~\ref{tab_eff}, it can be seen that our IDS performs consistently better in all test domains. In contrast to the other two augmentation strategies, i.e., MixStyle \cite{zhoudomain} and DSU \cite{li2022uncertainty}, it produces the performance gains of 3.6\% and 3.3\% in terms of average accuracy, indicating that such IDS augmentation strategy has a potential to synthesize more diverse domains. In fact, though both MixStyle and DSU are still able to reveal underlying domain information, the statistics augmented by IDS are significantly diverse, which suggests that it can create more inhomogeneous features. Overall, the ablation analyses in \tablename~\ref{tab_eff} verify the superiority of our suggested IDS against the other data augmentation schemes.
\begin{figure}[!t]
	\centering
	\subfigure[]{
		\includegraphics[width=0.4\linewidth]{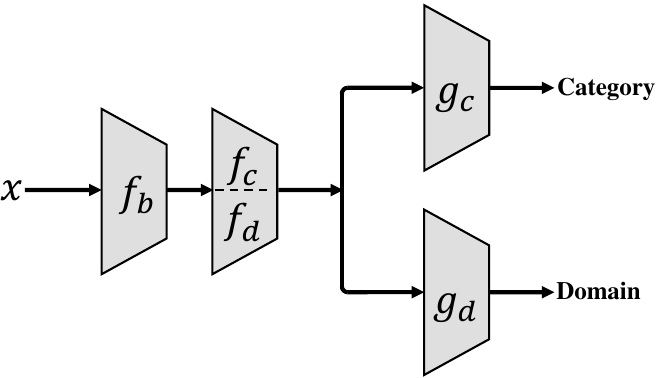}
		\label{fig:subfig_abs_branch_1}
	}
	\subfigure[]{
		\includegraphics[width=0.4\linewidth]{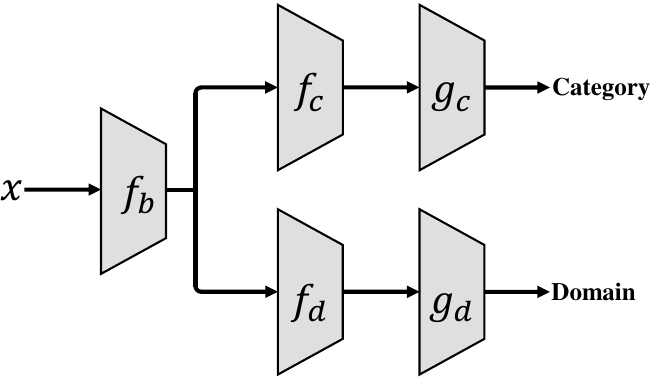}
		\label{fig:subfig_abs_branch_2}
	}
	\caption{A comparison of our Early-Forking Two-Branch framework with the classic Two-Branch Framework. \textbf{(a)} A Two-Branch Framework with shared feature extractors for $f_c$ and $f_d$. \textbf{(b)} An Early-Forking Two-Branch Framework. Here both $g_c$ and $g_d$ denote the final classifier in each branch.}
	\vspace{.1in}
	\label{fig_abs_branch}
\end{figure}

\textbf{Early-forking can effectively disentangle the casual/non-casual factors.}
As illustrated in \figurename~\ref{fig:subfig_abs_branch_1}, we design a two-branch framework with early-forking to disentangle the casual and non-casual features for DG-based activity recognition, which is built on the shared base feature extractor, with two attached lightweight classification heads having much fewer parameters \cite{atzmon2020causal,chen2021style,ganin2016domain}. In other words, our early-forking design is equivalent to applying a shared based feature extractor $f_b$ plus two separate frameworks, i.e., $F_c$ and $F_d$ for feature disentanglement. In fact, it has been previously stated that while removing such early-forking design from the two-branch framework as shown in \figurename~\ref{fig:subfig_abs_branch_2}, it would cause the causal feature extractor $F_c$ and the non-causal features $F_d$ to rely on each other too heavily, which might be very detrimental to the cross-domain activity recognition performance \cite{atzmon2020causal,chen2021style}. To support our hypothesis, we conduct further ablation experiments on the DSADS dataset to verify whether the proposed early-forking design is able to help disentangling the casual/non-casual factors in our two-branch framework for DG-based activity recognition. To ensure fair comparisons, we keep the other experimental settings unchanged. The compared results are listed in the 11th row of \tablename~\ref{tab_eff}. While removing such early-forking design from our two-branch framework, we find that it results in a dramatic accuracy decrease of round 5.4\%. The results agree well with our previous hypothesis that the proposed early-forking design can serve as an effective complement to the independence-based two-branch framework in disentangling the casual and non-casual features for domain-invariant representation learning.

\section{DISCUSSION}
\begin{figure}[]
	\centering
	\subfigure[ERM]{
		\includegraphics[width=0.40\linewidth]{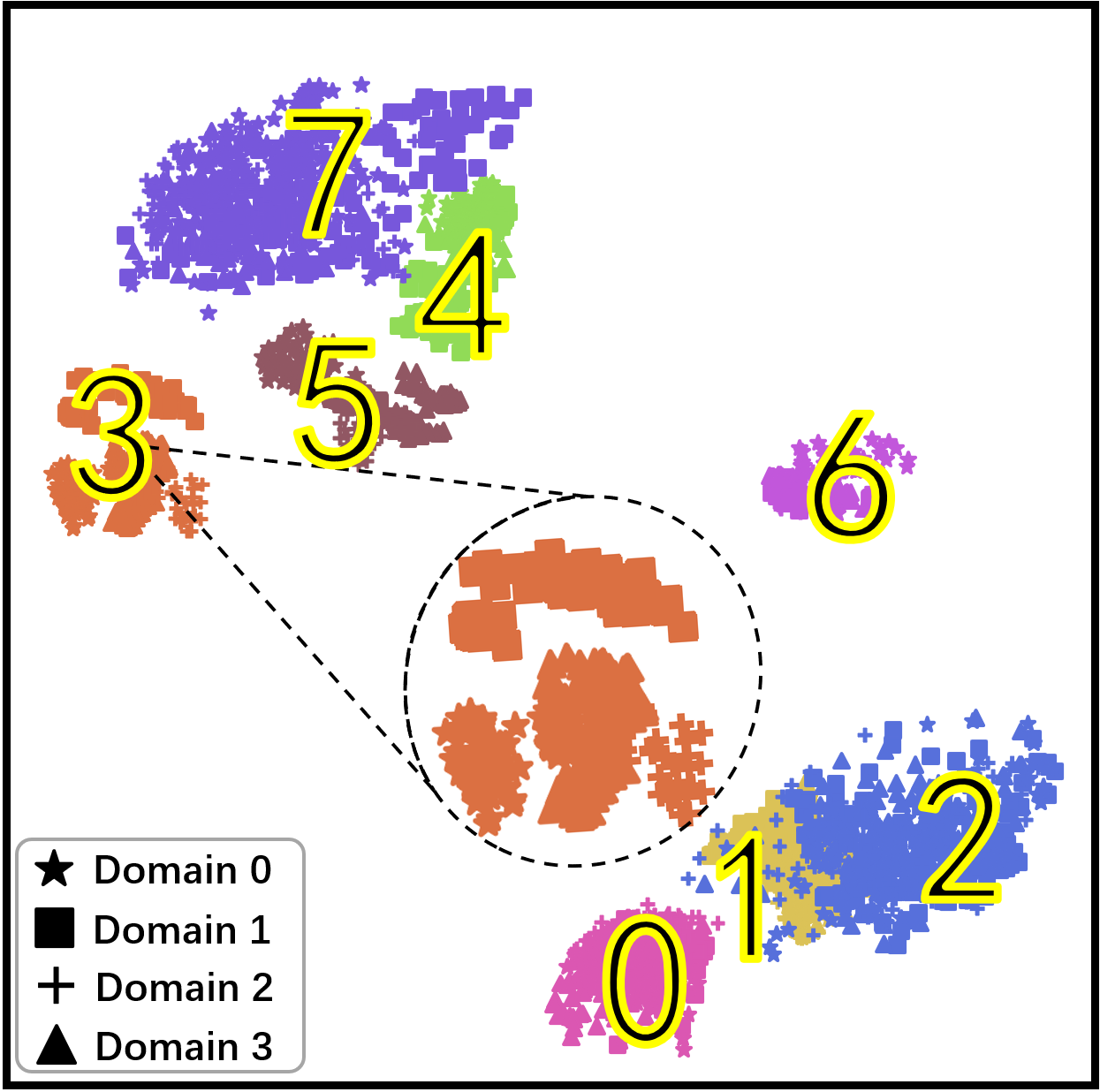}
		\label{fig:subfig_ts_erm}
	}
	\subfigure[Ours w/o CDPL \& IDS]{
		\includegraphics[width=0.40\linewidth]{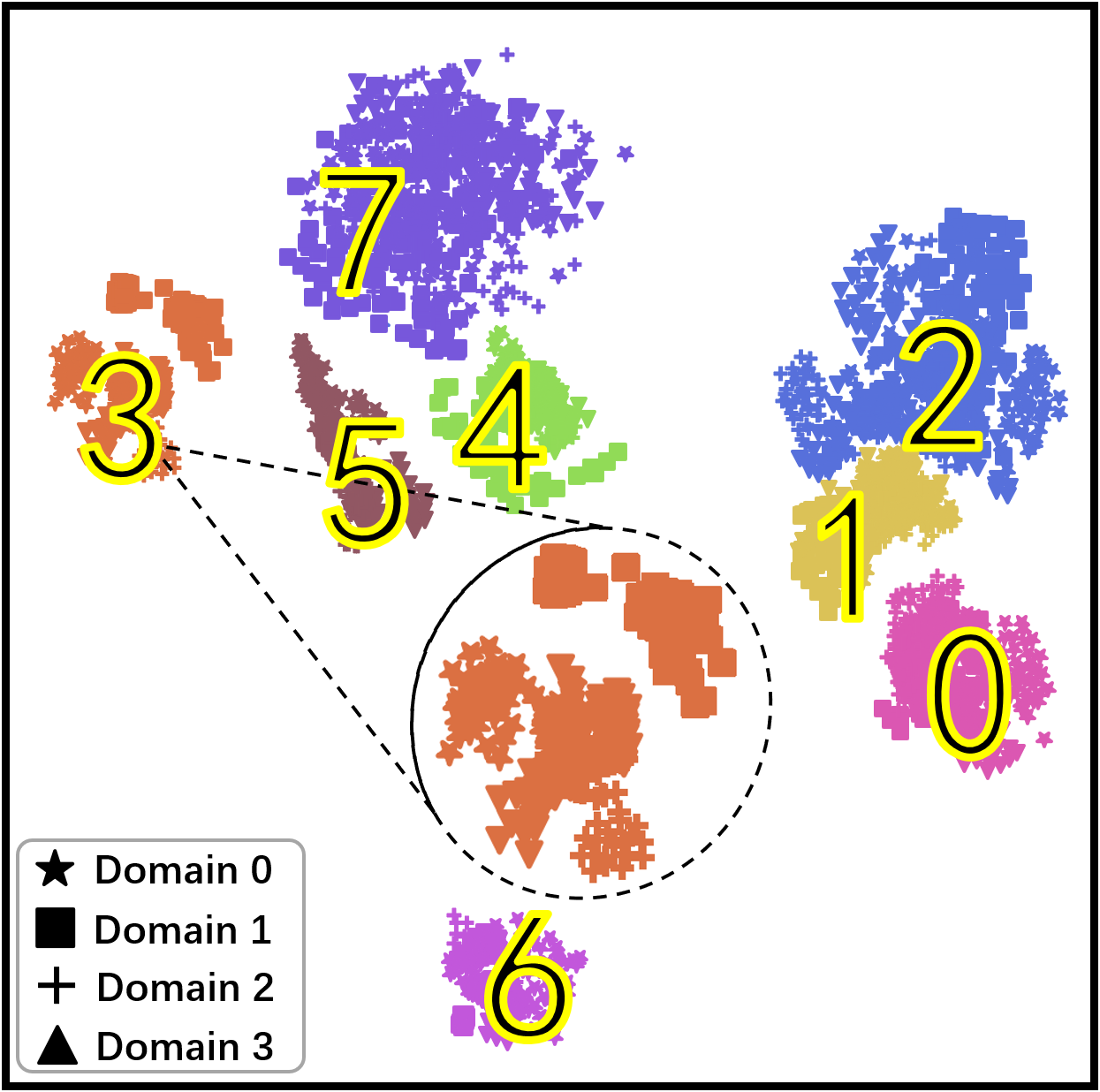}
		\label{fig:subfig_ts_hsic}
	} \\
	\subfigure[Ours w/o CDPL]{
		\includegraphics[width=0.40\linewidth]{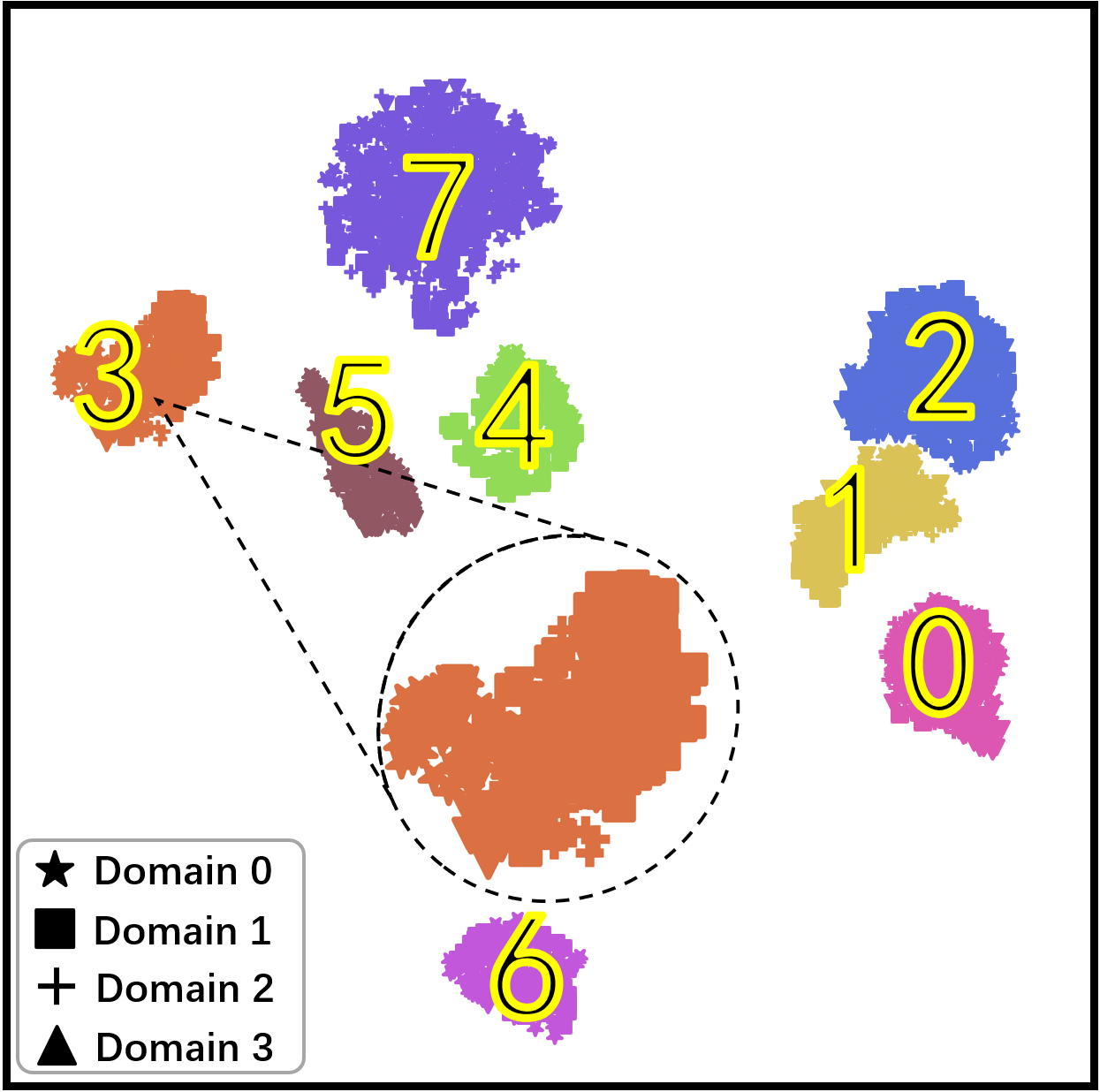}
		\label{fig:subfig_ts_rds}
	}
	\subfigure[Ours]{
		\includegraphics[width=0.40\linewidth]{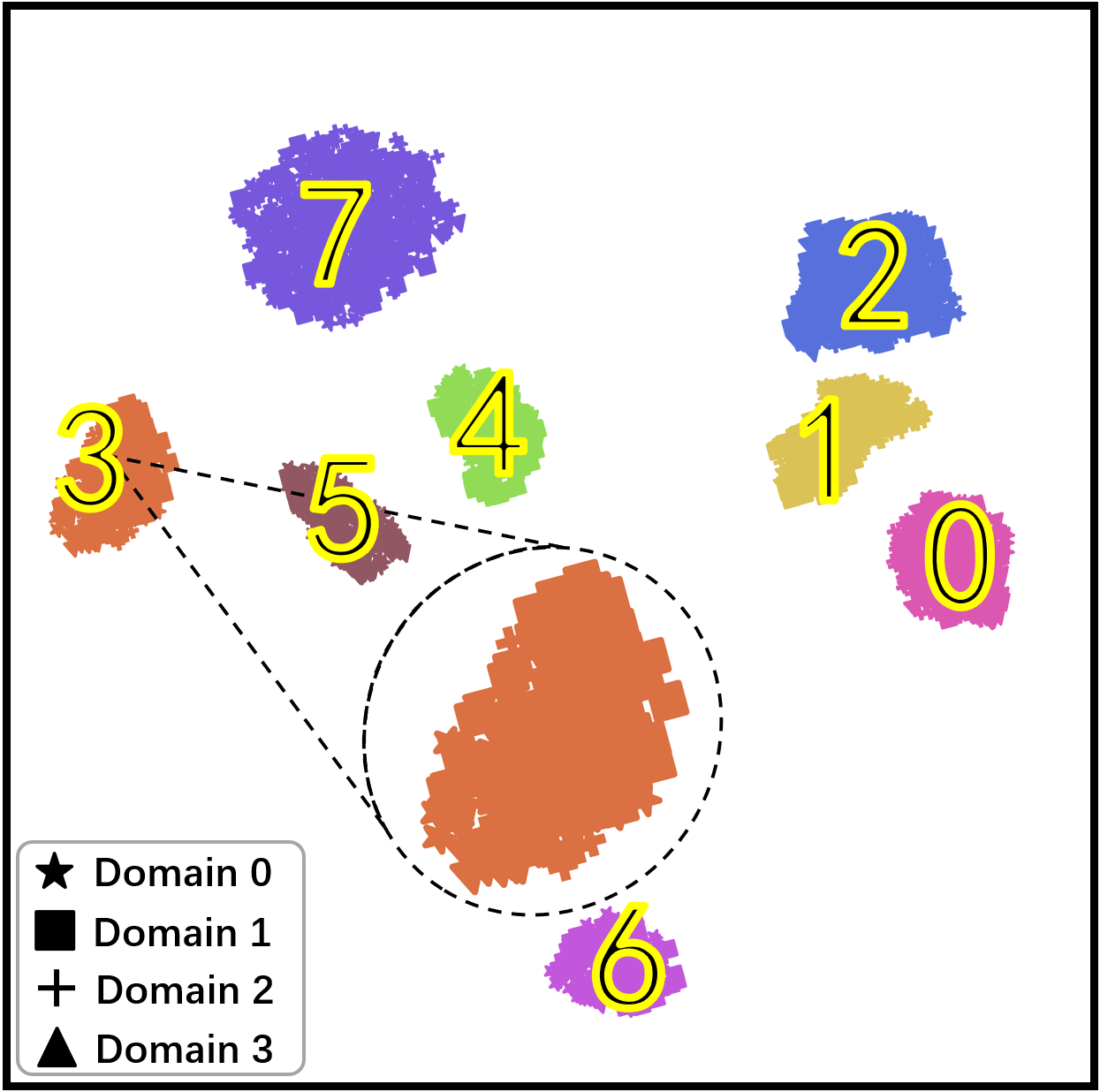}
		\label{fig:subfig_ts_ours}
	}
	\caption{Visualization of t-SNE embeddings of eight activity classes chosen from the DSADS dataset. Here, different colors represent different classes (0: "Descending stairs"; 1: "Standing"; 2: "Standing in an elevator still"; 3: "Running on a treadmill with a speed of 8 km/h"; 4: "Exercising on a stepper"; 5: "Exercising on a cross trainer"; 6: "Cycling on an exercise bike in horizontal"; 7: "Playing basketball"). Different shapes indicate different domains under cross-person setting. (a) It can be seen that the standard ERM baseline still generates multiple dispersed clusters (each class is represented by a color), but they are not well separated. (b) In contrast to (a), it can be seen that applying HSIC to our two-branch framework would enlarge the classification margin, which make multiple clusters be far from each other. However, their domains could not be well aligned. It can be seen that different domains denoted by shapes are still far from each other within every class; (c) In contrast to (b), it can be seen that after performing IDS, the domains represented by shapes are more domain-invariant, which are well mingled within every class. This is due to that IDS concentrates more on enhancing domain-invariant feature learning, which is able to generalize better across domains; (d) In contrast to (c), it can be seen that the whole version of our algorithm with CDPL further enhances such representation learning, which can not only force the domains to be more invariant, but also separate the classes better. Best viewed in color and zoom in.}
	\label{fig_ts}
\end{figure}

\begin{figure}[!t]
	\centering
	\subfigure[ERM]{
		\includegraphics[width=0.338\linewidth]{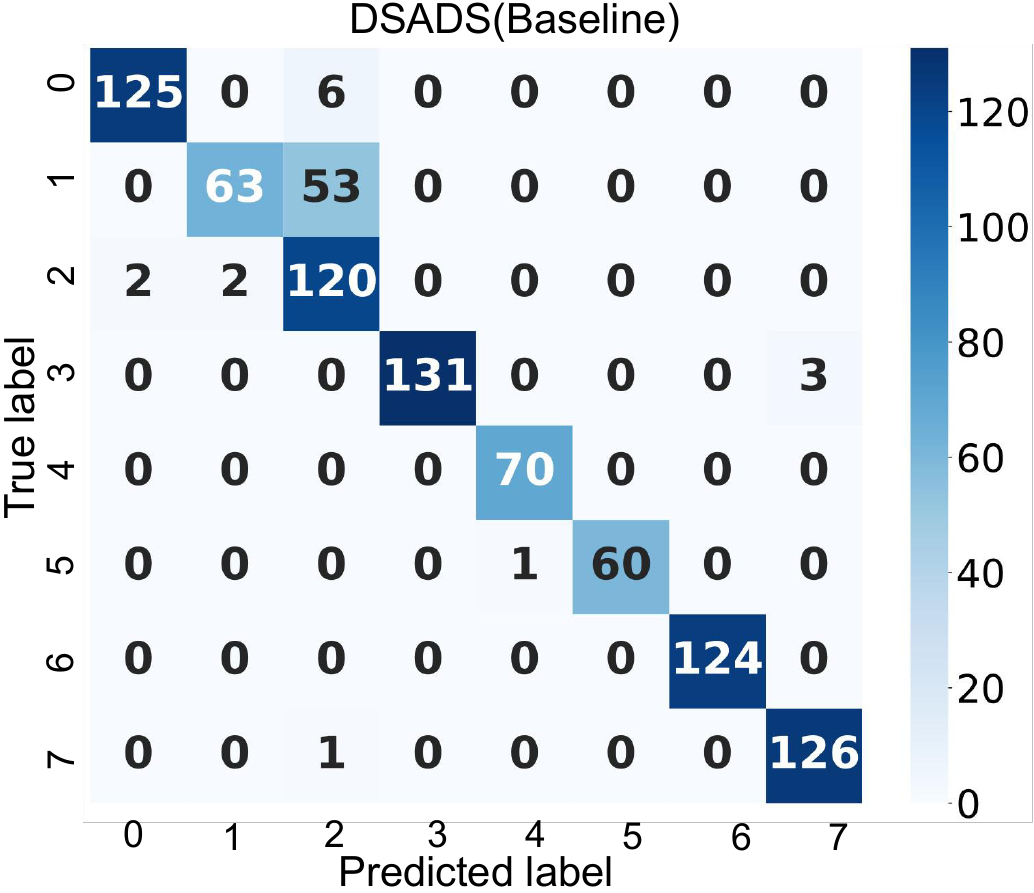}
		\label{fig:subfig_cf_1}
	}
	\subfigure[Ours]{
		\includegraphics[width=0.341\linewidth]{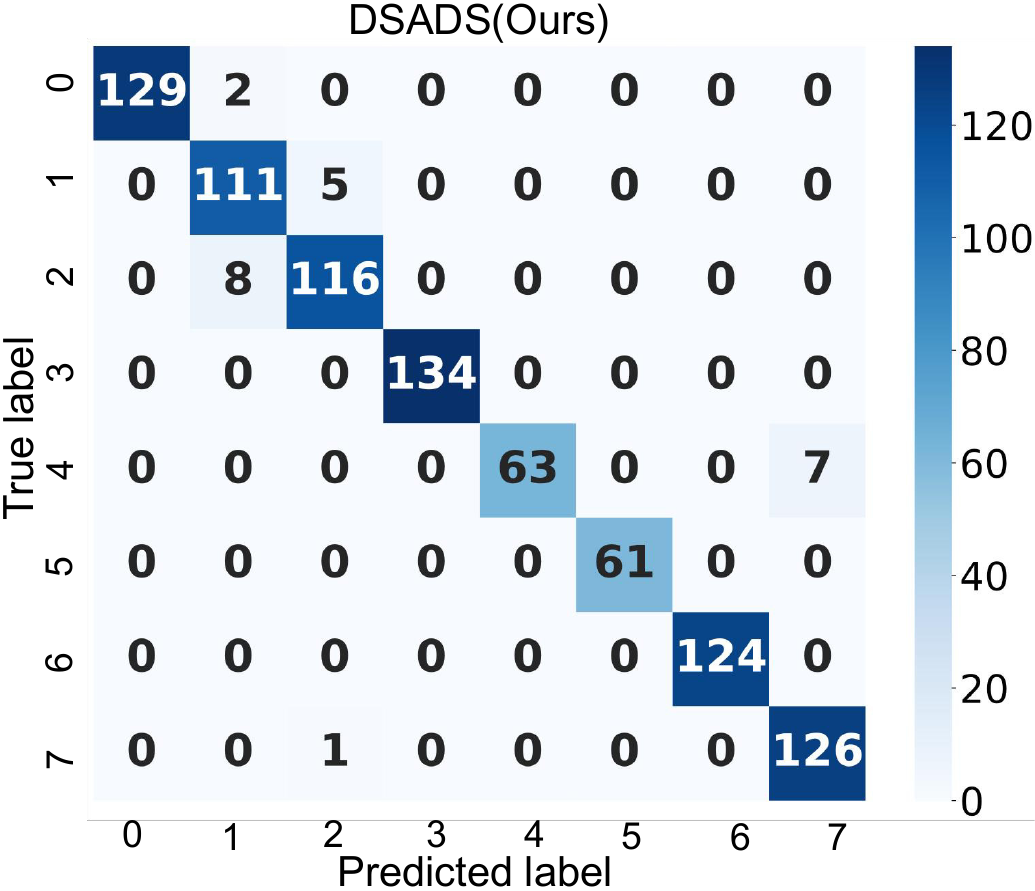}
		\label{fig:subfig_cf_2}
	}
	\caption{The confusion matrix illustrating the comparison between the proposed method and the ERM method on the DSADS dataset. The comparison between (a) and (b) indicates that our method significantly reduces the number of samples misclassified from "2" (indicating "Standing in an elevator still") to "1" (indicating "Standing").}
	\label{fig_cf}
\end{figure}

\subsection{Visualized Results}
\textbf{How to achieve effective disentanglement for categories and domains?} In order to provide an interpretable visualization analysis about how each component learns domain-related information and semantic feature, we select these activity samples from the DSADS dataset’s Target 0 in this cross-person scenario, and then apply the popular t-SNE technique to visualize their feature embeddings in a low-dimensional latent space, as illustrated in \figurename~\ref{fig_ts}. Similar quantitative classification results may also refer to our main ablation study in \tablename~\ref{tab_eff}. Intuitively, the better activity classification performance, the higher the clustering of points should be. We compare the complete version of our causality-inspired representation learning algorithm with its three variants, i.e., the standard ERM baseline, ERM+HSIC, and ERM+HSIC+IDS. We can make the following observations. 
First, as shown in \figurename~\ref{fig:subfig_ts_erm}, we can see that the standard ERM baseline still generates multiple dispersed clusters (each class is represented by a color), but they are not well separated, implying that that ERM still tends to learn entangled domain-specific information and semantic features. On the contrary, this is not the case for ours; Second, when comparing Figures \ref{fig:subfig_ts_erm} and \ref{fig:subfig_ts_hsic}, we can see that applying HSIC to our two-branch framework would further enlarge the classification margin to make the clusters be far from each other, which indicates that our causality-inspired HSIC can effectively concentrate on casual features (i.e., label-related activity semantic) to better separate classes, by disentangling the casual and non-casual features. However, we note that though different classes are well separated, the feature points from different domains are still separated. That is, their domains could not be well aligned. It can be seen that different domains denoted by shapes are still far from each other in every class, indicating that using HSIC alone in our two-branch design, still encodes some non-causal domain-specific information in the target features well; Third, when comparing  \figurename~\ref{fig:subfig_ts_erm} with \ref{fig:subfig_ts_rds}, we can see that after performing IDS, the domains represented by shapes are well mingled in every class. The result validates that IDS is able to further help disentangle the causal and noncausal features, which results in less domain-specific information and concentrates more on enhancing domain-invariant feature learning, generalizing better across domains; Finally, from \figurename~\ref{fig:subfig_ts_ours}, we can see that the whole version of our algorithm with CDPL can further enhance such representation learning, which not only forces the domains to be more invariant, but also improve activity classification results. The visualized analyses well support our hypothesis, which show that the whole version of our algorithm achieves the best results, where each module could play an essential role in feature disentanglement. In order to provide a quantitative analysis, we further compute the confusion matrices in \figurename~\ref{fig_cf} using the same settings in \figurename~\ref{fig_ts}. We focus on the category of incorrectly predicted activities. It can be seen that there exists a total of 53 samples belonging to ‘Standing’, which have been wrongly classified as ‘Standing in an elevator still’ by the standard ERM baseline. Our method can considerably decrease the number of misclassifications from 53 to 5, which again verifies the effectiveness of our causality-inspired learning approach.

\begin{figure}[!t]
	\centering
	\subfigure[Sensor positions]{
		\includegraphics[width=0.18\linewidth]{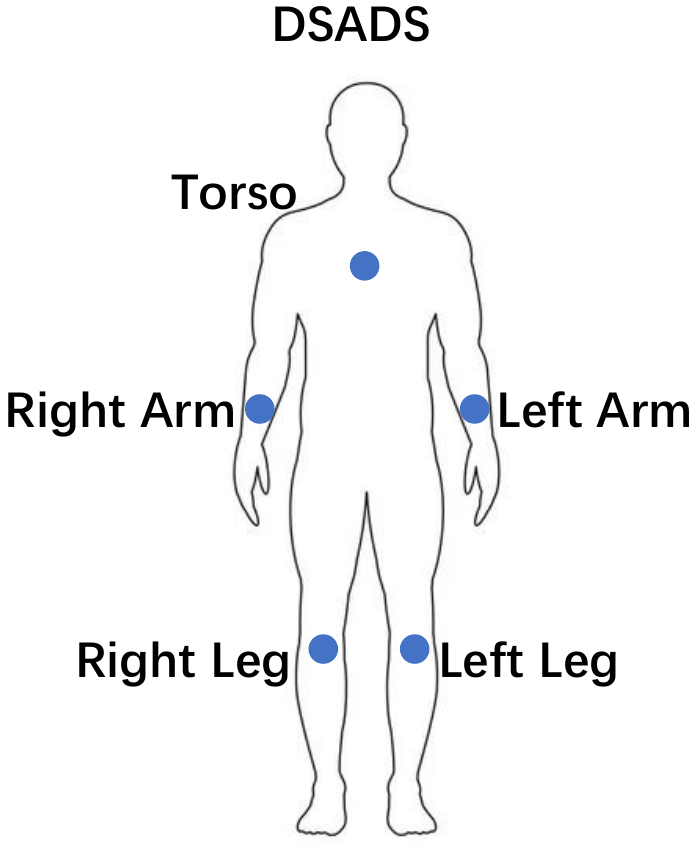}
		\label{fig:cross-position_a}
	}
	\subfigure[RA->RL]{
		\includegraphics[width=0.25\linewidth]{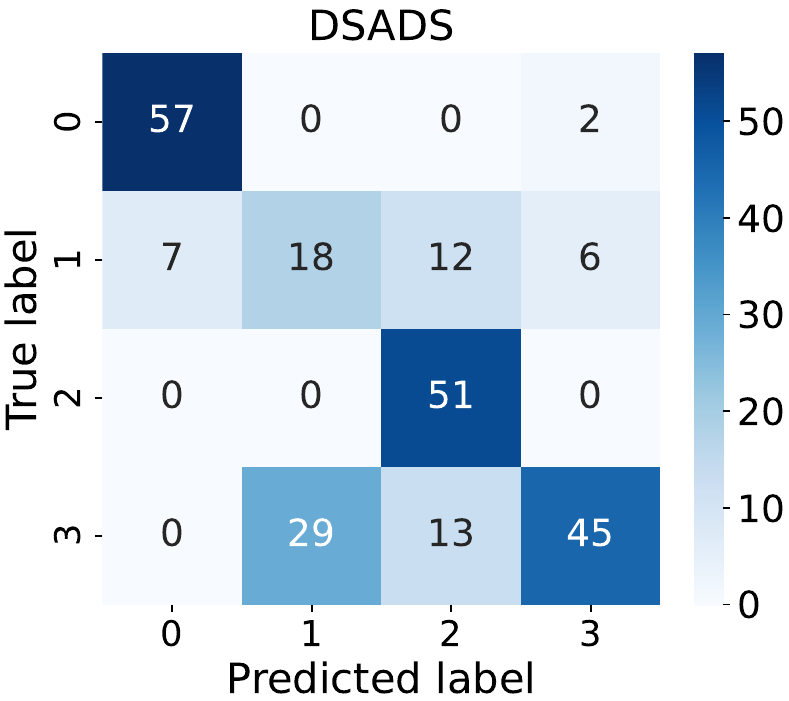}
		\label{fig:cross-position_b}
	}
	\subfigure[RA+LL->RL]{
		\includegraphics[width=0.25\linewidth]{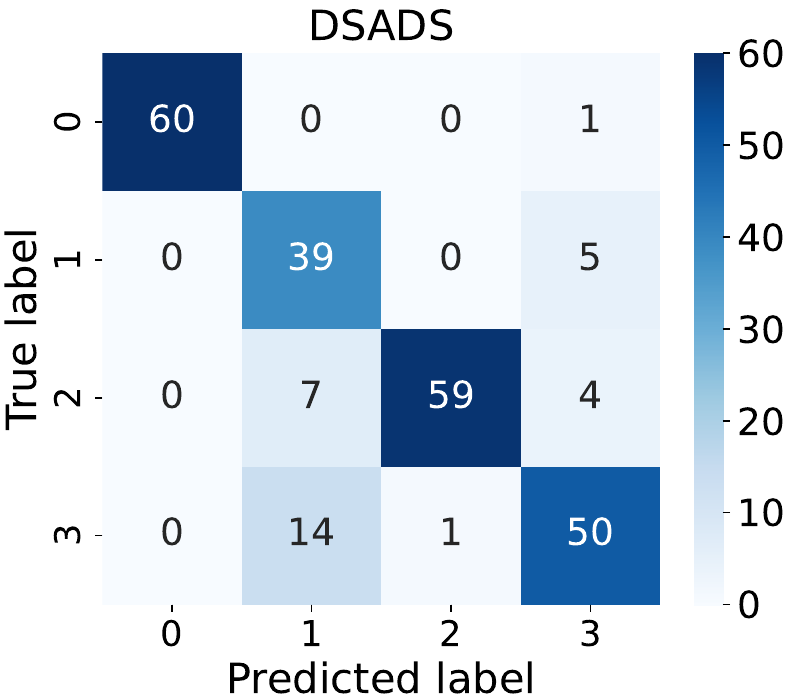}
		\label{fig:cross-position_c}
	}
	\subfigure[RA+LL+T->RL]{
		\includegraphics[width=0.25\linewidth]{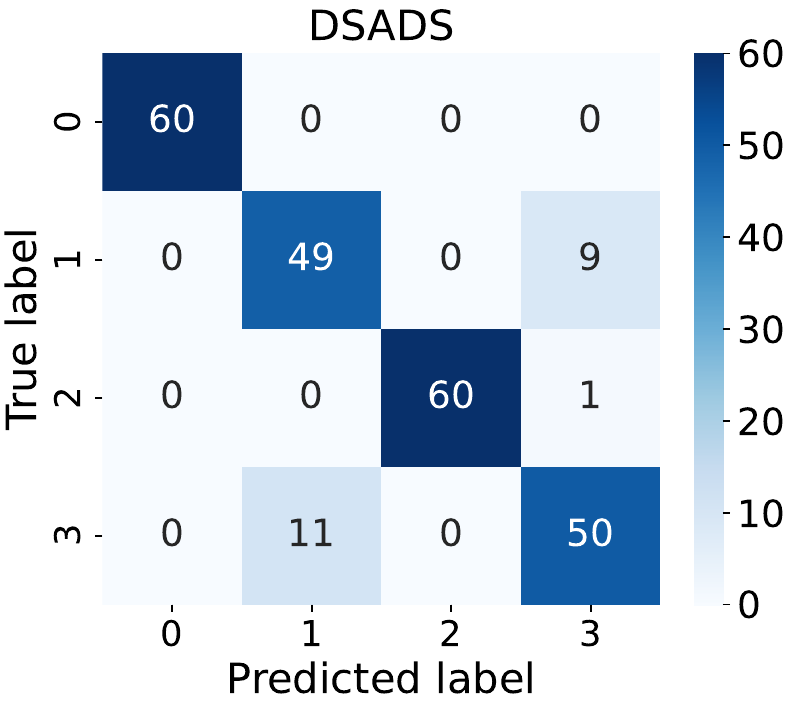}
		\label{fig:cross-position_d}
	}
	\vspace{-.10in}
	\caption{(a) A diagram showing the sensor locations on body parts in the DSADS dataset. (b) Right Arm sensor data as the source domain, and Right Leg sensor data as the target domain. (c) Right Arm and Left Leg sensor data as the source domain, and Right Leg sensor data as the target domain. (d) Right Arm, Left Leg, and Torso sensor data as the source domain, and Right Leg sensor data as the target domain. In (b), (c), and (d), 0,1,2, and 3 respectively denote "Ascending stairs", "Cycling on an exercise bike in a vertical position", "Jumping" and "Playing basketball".}
	\label{fig:cross-position}
\end{figure}
\subsection{Cross-Position Analysis}
\textbf{Is the similarity between different body parts is extremely crucial for cross-position activity recognition?} To facilitate our analysis, we present the detailed cross-position results of the same person on four exemplary activity categories, i.e., 0: “Ascending stairs”; 1: “Cycling on an exercise bike in a vertical position”; 2: “Jumping”; 3: “Playing basketball”. We take the right leg as our target domain. After fixing the target domain, we then select three different source domains to train our model by combing different body parts, e.g., Right Arm -> Right Leg, Right Arm + Left Leg -> Right Leg, and Right Arm + Left Leg + Torso -> Right Leg. In practice, such cross-position domain generalization is extremely challenging, since some body parts may be similar, or totally dissimilar without any shared body structure and action correlation. With regards to the most challenging activity “Cycling on an exercise bike in a vertical position”, we can make the following observations: Firstly, while only selecting the right arm as source domain , it can be seen that \figurename~\ref{fig:cross-position_b} exhibit the worst performance, which is because Right Arm is more dissimilar to Right Leg (LA). For instance, the low-body movement, e.g., “Cycling on an exercise bike in a vertical position” would be hard to detect by the IMU attached to the upper body; Secondly, while further adding the left leg as the source domain, the generalization performance will significantly increase, as illustrated in \figurename~\ref{fig:cross-position_c}. This is probably because that the most similar part to the right leg is its opposite (i.e., the left leg): Finally, adding torso as the source domain can further increase the results a little, as illustrated in \figurename~\ref{fig:cross-position_d}. Intuitively, since the Torso is more or less correlated with the other body parts, there may be more knowledge contained in multiple domains than a single domain. As shown in \tablename~\ref{tab_crpo}, for a different person, our method, as well as all existing DG-based algorithms produce the worst results because different people have exactly different body structure and moving patterns. There is still large room for performance improvement in such challenging cross-position research direction.

\begin{table*}[!t]
	\small
	\centering
	\caption{Average classification accuracy(\%) for different backbones on DSADS dataset under  cross-person setting. \textbf{Bold} means the best while \underline{underline} means the second-best.}
    \resizebox{1.0\textwidth}{!}{
	\begin{tabular}{cl|cccccccccc}
		\toprule
		&Target & $\text{ERM}$ & $\text{DANN}$ & $\text{CORAL}$ & $\text{Mixup}$ & $\text{GroupDRO}$ & $\text{RSC}$ & $\text{ANDMask}$ & FIXED&$\text{DIVERSIFY}$ & $\text{Ours}$ \\
		\midrule 
		\multirow{5}{*}{\rotatebox{90}{DSADS}}
		&CNNs & 80.3 & 85.6 & 85.4 & 87.0 & 85.9 & 82.9 & 81.4 & 84.8 &\underline{88.2}& \textbf{93.6} \\
		&ResNet & 82.8 & 86.6 & 84.8 & 87.4 & 86.2 & 83.1 & 82.2 & 85.6 &\underline{88.7}& \textbf{93.8}  \\
		&DeepConvLSTM & 79.6 & 81.1 & 82.5 & \underline{86.5} & 82.1 & 80.6 & 80.5 & 84.6 &86.3& \textbf{90.5}  \\
		&Transformer & 83.4 & 86.2 & 85.6 & 87.6 & 86.6 & 83.4 & 82.2 & 86.2 &\underline{90.0}& \textbf{94.1}  \\
		&AVG & 81.5 & 84.9 & 84.6 & 87.1 & 85.2 & 82.5 & 81.6 & 85.3 &\underline{88.3}& \textbf{93.0}  \\
		\bottomrule
	\end{tabular}
    }
	\vspace{-.25in}
	\label{tab_bf}
\end{table*}
\subsection{Backbone Flexibility}
\textbf{Can our method be extended to other model architectures?} As previously introduced, we have suggested an early-forking design in the two-branch framework. In order to further verify its universality or extensibility, we perform more evaluations on the DSADS dataset by applying it to different popular network backbones under cross-person evaluation setting. To be specific, we choose three widely-employed HAR backbone networks including ResNet \cite{he2016deep,huang2022channel}, DeepConvLSTM \cite{ordonez2016deep}, and Transformer \cite{vaswani2017attention} for our extensibility analysis. We conduct extensive comparisons with the other classical domain generalization baselines that are totally independent of backbone structures. To ensure fair comparisons, we keep other experimental settings the same as those in all three evaluations. The results are listed in \tablename~\ref{tab_bf}. 

We observe that our method consistently performs the best among all the cases, which significantly surpasses the second-best baseline such as DIVERSITY \cite{lu2024diversify} and Mixup \cite{zhang2018mixup} by a large margin. In contrast to smaller convolutional networks, it can be seen that our method can achieve better results on both ResNet and Transformers with an early-forking structure, which indicates that the suggested early-forking design can serve as a universal solution to the causality-inspired framework regardless of different network backbones, demonstrating its extensibility.

\begin{figure}[!t]
	\centering
	\includegraphics[width=0.8\linewidth]{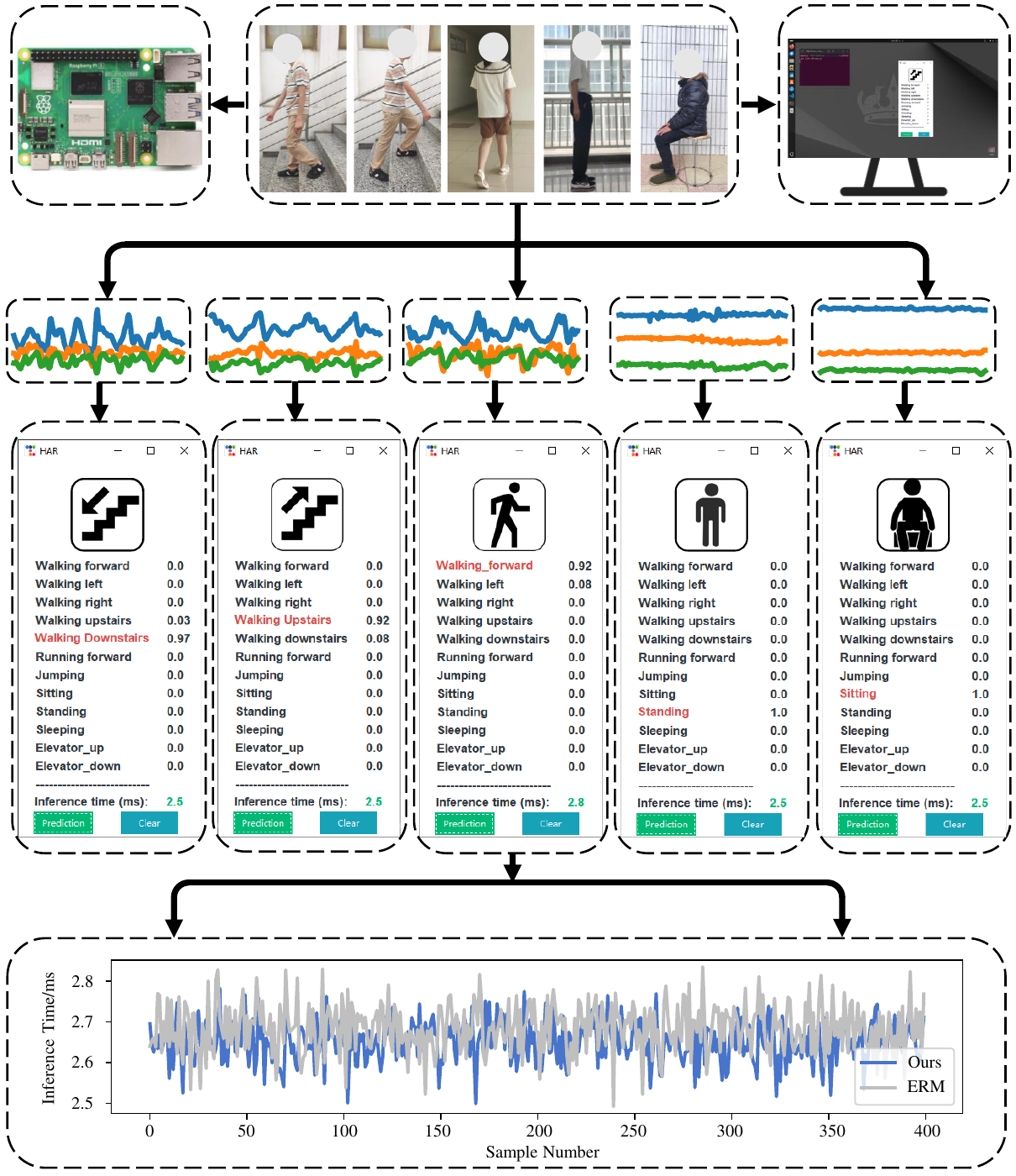}  
	\caption{The deployment of HAR models for people of different ages is demonstrated. The model trained on the USC-HAD dataset is deployed for practical HAR on a Raspberry Pi platform and validated for effectiveness across various activities such as "Walking Downstairs", "Walking Upstairs", "Walking Forward", "Standing", and "Sitting". The figure below compares the inference times between our ERM baseline method and our proposed method, showing that our approach does not compromise inference speed.}
	\vspace{-.1in}
	\label{fig_deploy}
\end{figure}

\begin{figure}[!t]
	\centering
	\subfigure[]{
		\includegraphics[width=0.31\linewidth]{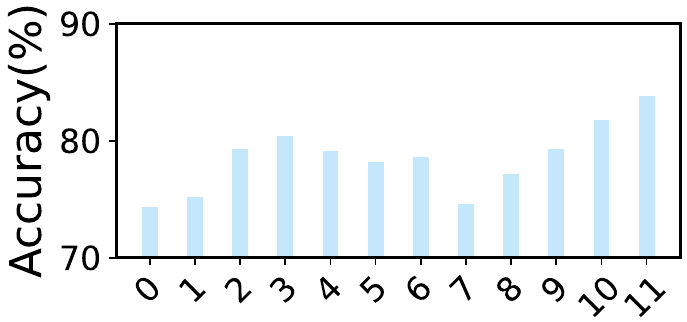}
		\label{fig_deploy_acc}
	}
	\subfigure[]{
		\includegraphics[width=0.31\linewidth]{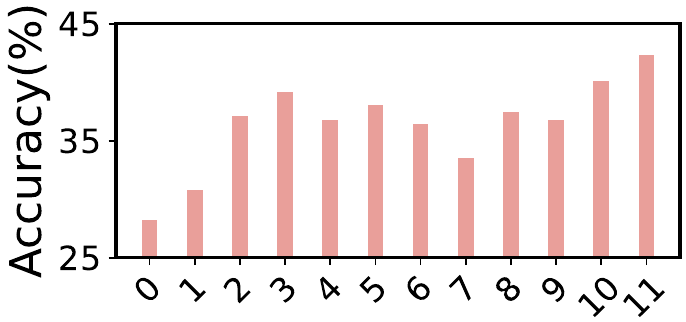}
		\label{fig_deploy_acc_cross_position}
	}
	\subfigure[]{
		\includegraphics[width=0.31\linewidth]{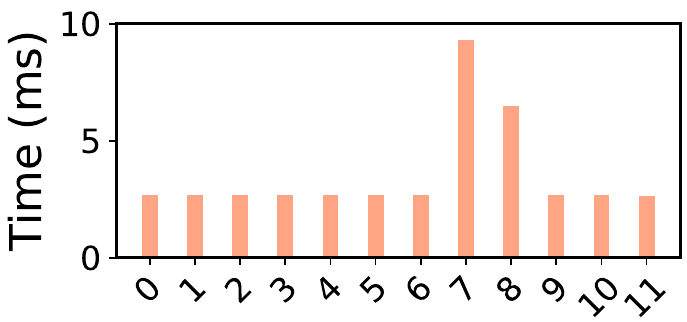}
		\label{fig_deploy_infer}
	}
	\vspace{-.2in}
	\caption{The average recognition accuracy and inference time for deploying HAR models are provided. "0", "1", "2", "3", "4", "5", "6", "7", "8", "9", "10", and "11" represent the methods: "ERM", "DANN", "CORAL", "Mixup", "GroupDRO", "RSC", "ANDMask", "GILE", "AdaRNN", "FIXED", "DIVERSIFY", and "Ours". \textbf{(a)} Our method achieves competitive average accuracy in activity recognition across people of different ages, compared to the other eleven methods. \textbf{(b)} Our method achieves competitive average accuracy in activity recognition across different body positions, compared to the other eleven methods. \textbf{(c)} A comparison of inference times between the eleven methods and our approach verifies the superiority of our method in terms of inference time.}
	\label{fig_deploy_acc_infer}
\end{figure}
\subsection{Real-World HAR Deployment and Inference Time} \label{infer}
\textbf{Can our method be effectively deployed in real-world scenarios?} To demonstrate the scalability of our causality-inspired learning approach, practical HAR deployment is conducted on an embedded Raspberry platform. Due to a good combability with PyTorch, the Raspberry Pi 5 Model B equipped with an ARM Cortex-A76 CPU has been selected for this purpose. The same version of the deep learning library, PyTorch 1.8.1, is installed on the Raspberry platform as that on the server, and a model for deployment trained on the USC-HAD dataset (e.g., $.pth$ files) is loaded for scalability evaluation. We conduct a practical case study with six subjects (i.e., two children, two adults, and two elderly people) ranging from 6 to 70 years old, each of whom wears the MotionNode at her/his front right hip. They are instructed to perform 12 types of activities belonging to the USC-HAD dataset. Sensors are connected to the Raspberry Pi device via serial port or Bluetooth. To ensure a fair comparison, here our implementation still shares the same experimental configurations as those in the USC-HAD dataset (e.g., sampling rate, window length, device location, and device type), meaning that the generated sensor readings could be collected when all the subjects perform the same activities using identical device locations and device types. It lasts about 2 minutes for a trial of each specific activity. \figurename~\ref{fig_deploy} illustrates five exemplar performed activities performed, their corresponding signal waveforms, as well as a main user interface written in Python. Generally speaking, the activity recognition models trained on adult data are likely to fail when inferring activities of children or elderly people. However, it is postulated that the HAR models trained on adult data via causal-inference learning can generalize effectively to children or elderly people’s data, despite diverse data distributions due to variations in their behavior styles or body shapes. As shown in \figurename~\ref{fig_deploy_acc}, although the target test data is not available for training (e.g., children and elderly people’s data), our method is still able to obtain the competitive average accuracy on unseen data from different age groups, indicating its good scalability under cross-person activity recognition scenario. In addition to the cross-person deployment with different user age groups, we further ask all the subjects to hold a Samsung Galaxy S24 phone in their right hands while performing the same category of activities. In this case, an Android application is also developed using Android Studio, which is in charge of reading data from the sensing devices and feeding it to the load model for activity inference. To align the tensor dimensions, we still adopt consistent sampling rates and window sizes. This is similar to the cross-dataset deployment, where the distribution of IMU data is clearly distinct from that of the USC-HAD dataset, due to different device locations and device types. We report averaged results in \figurename~\ref{fig_deploy_acc_cross_position}. Though our causality-inspired approach still demonstrates superiority over compared state-of-the-art algorithms, it produces the relatively low accuracy compared to the above cross-person results. In fact, comparing to cross-person scenario, such cross-dataset setting is extremely challenging, which simultaneously involves different user groups, device placements, device types, etc. Since there is still ample room for further performance improvement, these results highlight substantial necessity for sensor-based cross-dataset activity recognition in future study.

Moreover, in order to check whether the additional parameters brought by our two-branch framework during training would slow down actual inference speed, we measure the inference time of ERM, which may be viewed as a standard comparing baseline of our method, without incorporating extra domain augmentation and the independence constraint. A few popular DG-based baselines are also included for comparison purpose. As illustrated in the bottom panel of \figurename~\ref{fig_deploy}, we randomly test a total of 400 activity samples, by estimating their practical inference times elapsed, while a model begins to read a sample (e.g., "start time") and then return its prediction (e.g., "end time"). \figurename~\ref{fig_deploy_infer} summarizes the average accuracy and inference time. From the results, it can be seen that, despite minor fluctuations, our proposed method exhibits comparable or lower average inference times than other strong baselines while maintaining higher accuracy, thereby demonstrating its effectiveness and efficiency. In fact, although extra parameters have to be introduced into our two-branch framework for disentangling casual/non-casual factors, our approach still remains the same number of parameters as the standard ERM baseline, since the non-casual branch utilized during training stage would be discarded when deploying.

\section{Conclusion}
In this paper, a causality-inspired representation learning algorithm is proposed for sensor-based cross-domain activity recognition. Although time-series activity data across domains (e.g., different individuals or varying sensor positions of the same individual subject) exhibit diverse distributions, the causality-inspired learning approach seeks to address this challenge by focusing on the disentanglement of causal (i.e., activity semantics) and non-casual (i.e., domain) factors. The method is built upon a two-branch framework with early-forking design, where a shared base extractor is followed by two separate branches that are respectively responsible for learning causal and non-causal features. The HSIC measurement is then employed to disentangle both of features effectively. Moreover, an inhomogeneous domain sampling strategy is designed to enhance disentanglement capability, while CDPL is performed to prevent representation collapse. Extensive experiments conducted on four popular HAR benchmarks including DSADS, USC-HAD, PAMAP2, and UCI-HAR demonstrate that our method can considerably outperform most existing state-of-the-art DG-based baselines under various sensor-based domain generalization HAR settings. Detailed ablation studies and visualized results are proposed to reveal the underlying mechanism of casual inference, indicating its effectiveness, efficiency, and extensibility in sensor-based cross-domain activity recognition scenario. 

\textbf{Potential limitations and future directions.} In this paper, similar to a series of existing research efforts \cite{qian2021latent,lu2024diversify,lu2024fixed}, we mainly focus on generalization performance under distribution shifts, which assume multiple domains whose labels are known and static. However, such assumption could not always hold true in real-world scenario. One main property of time series data is its non-stationary property, implying that the statistical feature may dynamically change over time. For example, different persons may exhibit various patterns for the same activity, e.g., walking at different speeds. On the contrary, different persons may exhibit similar patterns, e.g., riding on even roads with the same speed. Moreover, the same individual even may show distinct patterns, e.g., riding on various uneven roads. There is no doubt that the model generalization performance will tend to severely deteriorate due to such non-stationery property. How to handle such dynamically varying distributions and learn out-of-distribution (OOD) representations for time series data remains more challenging and less explored. More importantly, in this area, most existing works \cite{qian2021latent,lu2024diversify,lu2024fixed} have typically assumed that both the training and test data share the same label space. Less attention has been paid to the case where test samples have non-overlapping labels (i.e., new classes) from training samples. Breaking this assumption, another related topic is the out-of-distribution (OOD) detection, which primarily concentrates on recognizing new unknown classes not encountered during training. Such open set activity classification tasks have rarely been explored from the perspective of domain generalization. Overall, we list the two potential limitations of our method in domain generalization, which will be left as our future work. Note that there is still large room for improvement in the suggested research directions. We hope that they may motivate future studies on generalizable sensor-based activity recognition problem.

\begin{acks}
This research is supported by the National Natural Science Foundation of China (Grant No. 62373194).
\end{acks}

\bibliographystyle{ACM-Reference-Format}
\bibliography{dx}

%
%
%
%
%
%
%
%

\end{document}